\documentclass[final]{siamltex}
\usepackage{graphicx} 
\usepackage{mathtools}
\usepackage{latexsym}
\usepackage[mathscr]{euscript}
\usepackage{xcolor}
\usepackage{algorithm,algorithmic}
\usepackage{fancyhdr}
\usepackage{epsfig}
\usepackage{mathptmx}       
\usepackage[most]{tcolorbox}
\usepackage{helvet}         
\usepackage{courier}        
\usepackage{type1cm}        
\usepackage{imakeidx}         
\usepackage[a4paper,textheight=23cm,textwidth=15cm]{geometry}
\usepackage{enumitem}
\usepackage{titlesec}
\usepackage{amsmath,amssymb,amsfonts}
\usepackage{hyperref}
\usepackage{tikz}
\usepackage{bm}
\usepackage{subcaption}
\usepackage{float}

\usepackage{booktabs}
 
\usepackage{multirow}
\usepackage{xcolor}
\usepackage{lineno}

\usetikzlibrary{arrows.meta,shapes,positioning,calc,fit,backgrounds,decorations.pathreplacing}
\usepackage{pifont}   

\newtheorem{remark}{remark}

\newcommand{\nd}{\noindent}

	\title{Dimensionality Reduction on Riemannian Manifolds in Data Analysis}
	\author{A. Elichi\thanks{Université du Littoral Cote d'Opale, LMPA, 50 rue F. Buisson, 62228 Calais-Cedex, France.} \and  K. Jbilou\footnotemark[1]} 
\date{}
	
\begin{document}
		\maketitle

        \begin{abstract}
            In this work, we investigate Riemannian geometry–based dimensionality reduction methods that respect the underlying manifold structure of the data. In particular, we focus on Principal Geodesic Analysis (PGA) as a nonlinear generalization of PCA for manifold-valued data, and extend discriminant analysis through Riemannian adaptations of other known dimensionality reduction methods. These approaches exploit geodesic distances, tangent space representations, and intrinsic statistical measures to achieve more faithful low-dimensional embeddings. We also discuss related manifold learning techniques and highlight their theoretical foundations and practical advantages. Experimental results on representative datasets demonstrate that Riemannian methods provide improved representation quality and classification performance compared to their Euclidean counterparts, especially for data constrained to curved spaces such as hyperspheres and symmetric positive definite manifolds. This study underscores the importance of geometry-aware dimensionality reduction in modern machine learning and data science applications.
        \end{abstract}
   
   \nd {\bf Keywords}: Riemannian geometry, dimensionality reduction, manifold learning, data analysis, structure-preserving algorithms, optimization on manifolds.
   
   \nd {\bf AMS} {65F99, 58B20, 68T09}
       
\section{Introduction}

Dimensionality reduction is a fundamental task in data analysis, machine learning, and pattern recognition, aiming to represent high-dimensional data in a lower-dimensional space while preserving its essential structure. Classical methods such as Principal Component Analysis (PCA) and Linear Discriminant Analysis (LDA) have been widely used due to their simplicity and solid theoretical foundations \cite{fukunaga2013introduction,jolliffe2002principal}. However, these techniques are inherently Euclidean and rely on linear assumptions, which may be inadequate when data exhibit nonlinear structures or are constrained to non-Euclidean spaces.
\\
In many modern applications, including computer vision, signal processing, medical imaging, and shape analysis, data naturally reside on nonlinear manifolds rather than in flat Euclidean spaces. Examples include normalized feature vectors lying on hyperspheres, covariance descriptors modeled as symmetric positive definite (SPD) matrices, and shape spaces with intrinsic geometric constraints \cite{arsigny2007geometric,pennec2006intrinsic}. Ignoring the underlying geometry in such cases can lead to distorted representations and suboptimal performance.
\\
To address these limitations, Riemannian geometry–based dimensionality reduction methods have gained increasing attention. By explicitly modeling the manifold structure of the data, these methods exploit intrinsic notions such as geodesic distances, tangent spaces, and Riemannian metrics. One prominent example is Principal Geodesic Analysis (PGA), introduced as a nonlinear generalization of PCA for manifold-valued data \cite{fletcher2004principal}. PGA captures the main modes of variation along geodesics, providing a geometry-aware alternative to linear projections.
\\
Beyond unsupervised learning, supervised dimensionality reduction techniques have also been extended to the Riemannian setting. Riemannian versions of discriminant analysis adapt classical criteria to manifold-valued data by replacing Euclidean distances and statistics with their intrinsic counterparts \cite{harandi2014manifold,pennec2019riemannian}. These approaches have demonstrated improved classification performance, particularly when data lie on curved spaces such as hyperspheres or SPD manifolds.
\\
In parallel, a wide range of manifold learning methods including Isomap, Locally Linear Embedding (LLE), and Laplacian Eigenmaps—have been proposed to uncover low-dimensional structures in nonlinear data \cite{roweis2000nonlinear,tenenbaum2000global}. While effective in many scenarios, these techniques are often extrinsic and may lack a clear statistical interpretation on Riemannian manifolds.
\\
In this work, we investigate dimensionality reduction methods grounded in Riemannian geometry, with a particular emphasis on PGA and Riemannian adaptations of classical discriminant and projection-based techniques. By leveraging intrinsic geometric tools, the proposed framework yields low-dimensional embeddings that better respect the data’s manifold structure. Experimental evaluations on representative datasets demonstrate that Riemannian methods consistently outperform their Euclidean counterparts in terms of representation fidelity and classification accuracy. These results highlight the importance of geometry-aware dimensionality reduction in modern machine learning and data science.\\

The paper is organized as follows.
Section~2 introduces the mathematical preliminaries of Riemannian geometry,
including smooth manifolds, tangent spaces, Riemannian metrics, and geodesic
distance, as well as important examples such as the Grassmann, Stiefel, and
SPD manifolds.
Section~3 reviews optimization on Riemannian manifolds and presents fundamental
concepts such as the Riemannian gradient, retractions, and convergence guarantees
for first-order methods.
Section~4 describes Principal Geodesic Analysis (PGA) as a natural extension of
PCA to manifold-valued data.
Section~5 introduces Riemannian Robust Principal Component Analysis (RRPCA) for
handling outliers in manifold-valued datasets.
Section~6 extends Orthogonal Neighborhood Preserving Projections (ONPP) to
Riemannian manifolds, with specializations to SPD and Grassmann manifolds.
Section~7 presents Riemannian Laplacian Eigenmaps for nonlinear dimensionality
reduction on manifolds. Section~8 discusses extensions to supervised learning, including Linear
Discriminant Analysis on Riemannian manifolds while Section 9 introduces Riemannian Isomap method and in Section 10 we give a description of  Riemannian Support Vector Machine (RSVM). We end this work by some numerical experiments.

	\section{Preliminaries and definitions}	
		In this section, we give some mathematical material related to the Riemannian geometry to be used in the paper, \cite{Jbilou2026}.

        \medskip
	\begin{definition}
		A Riemannian manifold $\mathcal{M}$  is a topological space such that each element of $ \mathcal{M}$ has a neighborhood homeomorphic to $\mathbb{R}^d$ such that
        \begin{enumerate}
            \item Any two distinct points have disjoint neighborhoods (Hausdorff). 
            \item Second-countable: There exists a countable basis for the topology of $\mathcal{M}$. 
            \item {Locally Euclidean.}
For every point $p \in \mathcal{M}$, there exists an open neighborhood
$U \subset \mathcal{M}$ and a homeomorphism
\[
\varphi : U \to \varphi(U) \subset \mathbb{R}^n .
\]
The collection of such pairs $(U,\varphi)$ forms an \emph{atlas} of
$\mathcal{M}$.
Additionally, the atlas is required to be \emph{smooth}, meaning that for
any two overlapping charts $(U,\varphi)$ and $(V,\psi)$, the transition map
\[
\psi \circ \varphi^{-1} :
\varphi(U \cap V) \to \psi(U \cap V)
\]
is infinitely differentiable ($C^\infty$).

            \item Equipped with a smoothly varying inner product $g_p$ on each tangent space associated to the point $p$ of $\mathcal{M}$.
            \end{enumerate}
        \end{definition}

        \medskip
	\nd 	{Examples:}
		\begin{itemize}
			\item Sphere: $\mathbb{S}^n = \{ x \in \mathbb{R}^{n+1} \mid \|x\| = 1 \}$ 
			\item Symmetric positive definite (SPD) matrices: $\mathcal{S}_{++}^d = \{ C \in \mathbb{R}^{d \times d} \mid C = C^\top, \, x^\top C x > 0, \forall x \neq 0\}$  
			\item Grassmann manifold:  $\mathrm{Gr}(p,n)$, the space of $p$-dimensional subspaces of $\mathbb{R}^n$.
		\end{itemize}
		
	\medskip
		
		\begin{definition}[Tangent Spaces]
		The {tangent space} at point $p \in \mathcal{M}$, denoted $T_p\mathcal{M}$, is the set of tangent vectors at $p$:
		\[
		T_p \mathcal{M} = \{ v \in \mathbb{R}^n \mid v = \dot{\gamma}(0), \text{ for some curve } \gamma(t) \text{ with } \gamma(0) = p \}
		\]
        \end{definition}
		
\medskip

		\begin{definition}[Riemannian Metric]
		A {Riemannian metric} $g$ assigns an inner product on each tangent space:
		\[
		g_p : T_p \mathcal{M} \times T_p \mathcal{M} \to \mathbb{R}, \quad g_p(u,v) = \langle u, v \rangle_p
		\]
		- Length of a curve $\gamma : [0,1] \to  \mathcal{M}$:
		\[
		L(\gamma) = \int_0^1 \|\dot{\gamma}(t)\|_{\gamma(t)} \, dt=\int_0^1 \sqrt{ g_{\gamma(t)}(\dot{\gamma}(t), \dot{\gamma}(t)) } dt
		\]
		- Geodesic distance:
		\[
		d_\mathcal{M}(p,q) = \inf_\gamma \, L(\gamma)
		\]
\end{definition}

\subsection{Grassmann Manifold}

The \emph{Grassmann manifold}, denoted by $\mathrm{Gr}(p,n)$, is the set of all $p$-dimensional linear subspaces of $\mathbb{R}^n$:
\[
\mathrm{Gr}(p,n)
=
\left\{
\mathcal{U} \subset \mathbb{R}^n
\;\middle|\;
\dim(\mathcal{U}) = p
\right\}.
\]
Each point on $\mathrm{Gr}(p,n)$ represents an equivalence class of matrices whose columns span the same subspace.

\nd {Matrix Representation:} 
A point $\mathcal{U} \in \mathrm{Gr}(p,n)$ can be represented by an orthonormal basis matrix
\[
U \in \mathbb{R}^{n \times p},
\qquad
U^\top U = I_p,
\]
where different matrices $U$ and $UQ$, with $Q \in \mathrm{O}(p)$, represent the same point on the Grassmann manifold:
\[
\mathcal{U} = \mathrm{span}(U) = \mathrm{span}(UQ).
\]
The Grassmann manifold can be expressed as a quotient space
\[
\mathrm{Gr}(p,n)
=
\mathrm{St}(p,n) / \mathrm{O}(p),
\]
where $\mathrm{St}(p,n) = \{ U \in \mathbb{R}^{n \times p} \mid U^\top U = I_p \}$ is the Stiefel manifold and $\mathrm{O}(p)$ is the orthogonal group.
\\
The tangent space at a point $U \in \mathrm{Gr}(p,n)$ is given by
\[
T_U \mathrm{Gr}(p,n)
=
\left\{
Z \in \mathbb{R}^{n \times p}
\;\middle|\;
U^\top Z = 0
\right\}.
\]
\\
The canonical Riemannian metric on $\mathrm{Gr}(p,n)$ is defined as
\[
\langle Z_1, Z_2 \rangle
=
\mathrm{tr}(Z_1^\top Z_2),
\quad
Z_1, Z_2 \in T_U \mathrm{Gr}(p,n).
\]
Given two subspaces $\mathcal{U}_1, \mathcal{U}_2 \in \mathrm{Gr}(p,n)$ represented by $U_1$ and $U_2$, let
\[
U_1^\top U_2 = Q \cos(\Theta) R^\top
\]
be the compact SVD, where $\Theta = \mathrm{diag}(\theta_1,\dots,\theta_p)$ are the principal angles.
The geodesic distance is
\[
d_{\mathrm{Gr}}(\mathcal{U}_1, \mathcal{U}_2)
=
\left( \sum_{i=1}^p \theta_i^2 \right)^{1/2}.
\]
The exponential and logarithmic maps on $\mathrm{Gr}(p,n)$ admit closed-form expressions in terms of matrix exponentials and SVDs, enabling efficient optimization and learning algorithms on the manifold.

\begin{remark}
The Grassmann manifold naturally arises in applications involving subspace learning, dimensionality reduction, computer vision, and signal processing.
It is invariant to changes of basis and captures only the underlying subspace structure.
\end{remark}

\subsection{Stiefel Manifold}

The \emph{Stiefel manifold}, denoted by $\mathrm{St}(p,n)$, is the set of all ordered orthonormal $p$-frames in $\mathbb{R}^n$, i.e., all matrices with orthonormal columns:
\[
\mathrm{St}(p,n)
=
\left\{
U \in \mathbb{R}^{n \times p}
\;\middle|\;
U^\top U = I_p
\right\}.
\]
Each point on $\mathrm{St}(p,n)$ represents a collection of $p$ orthonormal vectors in $\mathbb{R}^n$.\\
The Stiefel manifold differs from the Grassmann manifold in that it preserves the \emph{ordering and orientation} of the basis vectors.
Two matrices $U_1, U_2 \in \mathrm{St}(p,n)$ represent the same point on the Grassmann manifold $\mathrm{Gr}(p,n)$ if and only if
\[
U_2 = U_1 Q, \quad Q \in \mathrm{O}(p),
\]
where $\mathrm{O}(p)$ is the orthogonal group.
The tangent space at a point $U \in \mathrm{St}(p,n)$ is given by
\[
T_U \mathrm{St}(p,n)
=
\left\{
Z \in \mathbb{R}^{n \times p}
\;\middle|\;
U^\top Z + Z^\top U = 0
\right\}.
\]
Any tangent vector can be decomposed as
\[
Z = U A + U_\perp B,
\]
where $A \in \mathbb{R}^{p \times p}$ is skew-symmetric and $B \in \mathbb{R}^{(n-p) \times p}$ is arbitrary.
A commonly used Riemannian metric on $\mathrm{St}(p,n)$ is the \emph{canonical metric}:
\[
\langle Z_1, Z_2 \rangle
=
\mathrm{tr}\!\left(
Z_1^\top
\left(
I - \tfrac{1}{2} U U^\top
\right)
Z_2
\right),
\quad
Z_1, Z_2 \in T_U \mathrm{St}(p,n).
\]
Closed-form expressions for geodesics on $\mathrm{St}(p,n)$ exist but involve matrix exponentials.
In practice, retractions such as the QR-based retraction are often used: $U+Z=QR$ and 
\[
\mathrm{Retr}_U(Z) = \mathrm{qf}(U + Z)=Q,
\]
where $\mathrm{qf}(\cdot)$ denotes the orthonormal factor of the QR decomposition.
\\
The Stiefel manifold appears naturally in problems with orthogonality constraints, including
principal component analysis, subspace tracking, independent component analysis,
and optimization problems with orthonormality constraints.

\paragraph{Remarks.}
Unlike the Grassmann manifold, the Stiefel manifold distinguishes between different orthonormal bases of the same subspace.
When only the subspace matters, the Grassmann manifold provides a more natural representation.

\begin{table}[h!]
\centering
\caption{Glossary of Riemannian Manifold Terms}
\begin{tabular}{l|p{9cm}}
\hline
\textbf{Term} & \textbf{Definition} \\
\hline
Manifold & A space that locally resembles Euclidean space $\mathbb{R}^n$ but can have a curved global structure. \\
\hline
Riemannian Manifold & A manifold equipped with a metric that defines distances and angles locally. \\
\hline
Tangent Space & A linear space attached to a point on the manifold, where standard linear algebra can be applied. \\
\hline
Logarithmic Map & "Generalization of \textbf{subtraction} in Riemannian space: $\text{Log}_{x_m}(x_n) = \xi$, point on $\mathcal{M}$ $\times$ point on $\mathcal{M}$ $\to$ tangent vector"
 \\
\hline
Exponential Map & "Generalization of \textbf{addition} in Riemannian space: $\text{Exp}_{x_m}(\xi) = x_n$, point on $\mathcal{M}$ $\times$ tangent vector $\to$ point on $\mathcal{M}$".
\\
\hline
Geodesic &\textbf{Locally minimizing curves} between two points on the manifold or The locally  "straightest" or shortest path between two points on the manifold.\\
\hline
Fréchet Mean / Karcher Mean & The point on the manifold that minimizes the sum of squared geodesic distances to all data points. \\
\hline
Principal Geodesic Analysis (PGA) & Extension of PCA to manifolds, identifying directions of maximum variance in the tangent space. \\
\hline
Riemannian Gradient & Generalization of the classical gradient that respects the manifold's geometry. \\
\hline
Riemannian Optimization & Optimization of functions defined on manifolds. \\
\hline
Stiefel Manifold & The set of $n \times p$ matrices with orthonormal columns. \\
\hline
Grassmann Manifold & The set of all $p$-dimensional linear subspaces of $\mathbb{R}^n$. \\
\hline
SPD Manifold & The set of symmetric positive definite (SPD) matrices. \\
\hline
Riemannian Distance & The shortest distance along a geodesic between two points on the manifold. \\
\hline
Projection onto Tangent Space & Mapping points to the tangent space using the logarithmic map for linear computations. \\
\hline
\end{tabular}
\label{tab:riemannian-glossary}
\end{table}


\begin{figure}[H]
    \centering
    \begin{subfigure}[t]{0.48\textwidth}
        \centering
        \includegraphics[width=\textwidth]{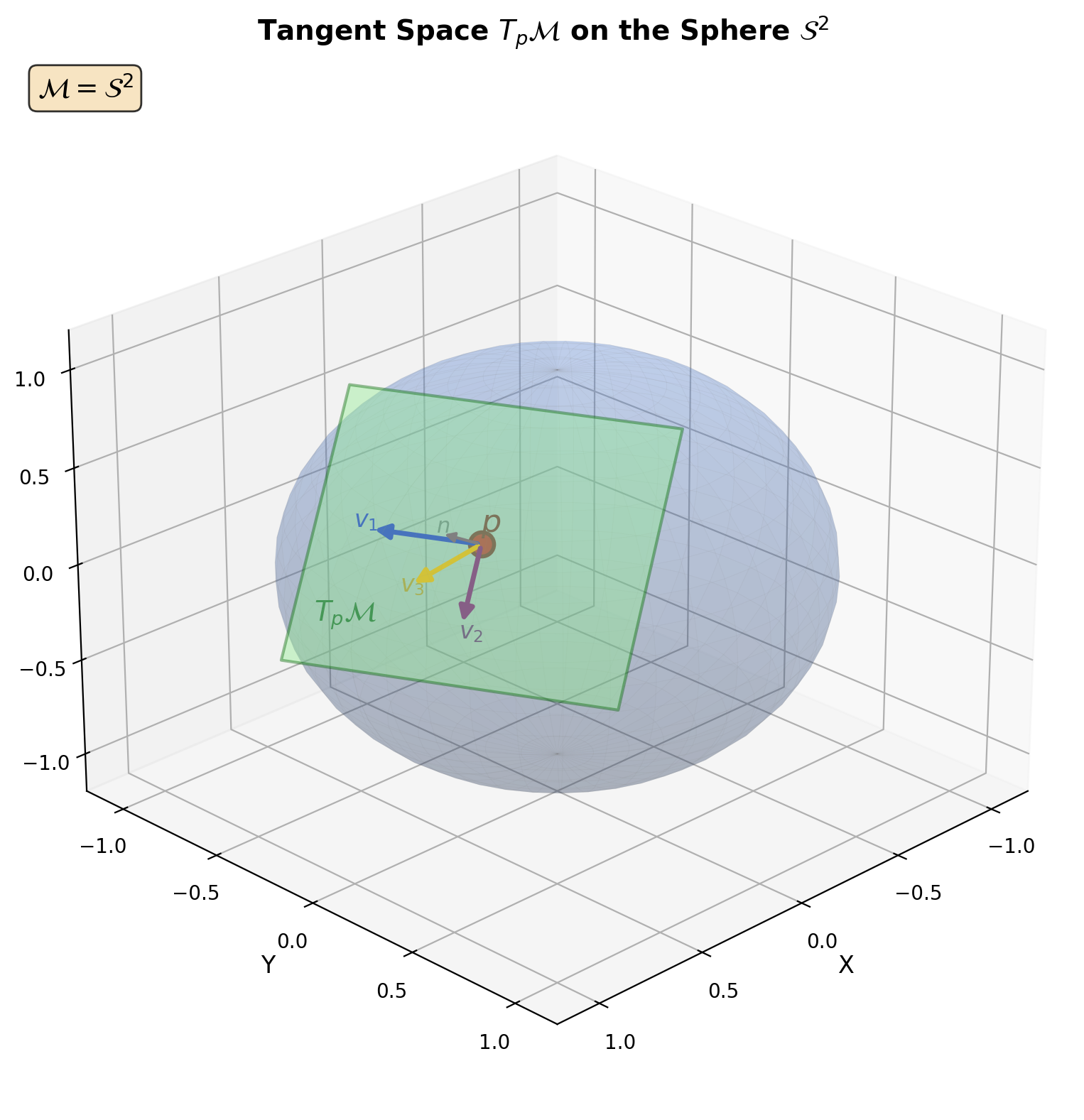}
        \caption{Tangent space $T_p\mathcal{M}$ on $\mathbb{S}^2$. The tangent plane at $p$ contains all vectors orthogonal to the normal $n=p$. Vectors $v_1, v_2, v_3 \in T_p\mathcal{M}$ represent infinitesimal directions on the manifold.}
        \label{fig:tangent_space_basic}
    \end{subfigure}
    \hfill
    \begin{subfigure}[t]{0.48\textwidth}
        \centering
        \includegraphics[width=\textwidth]{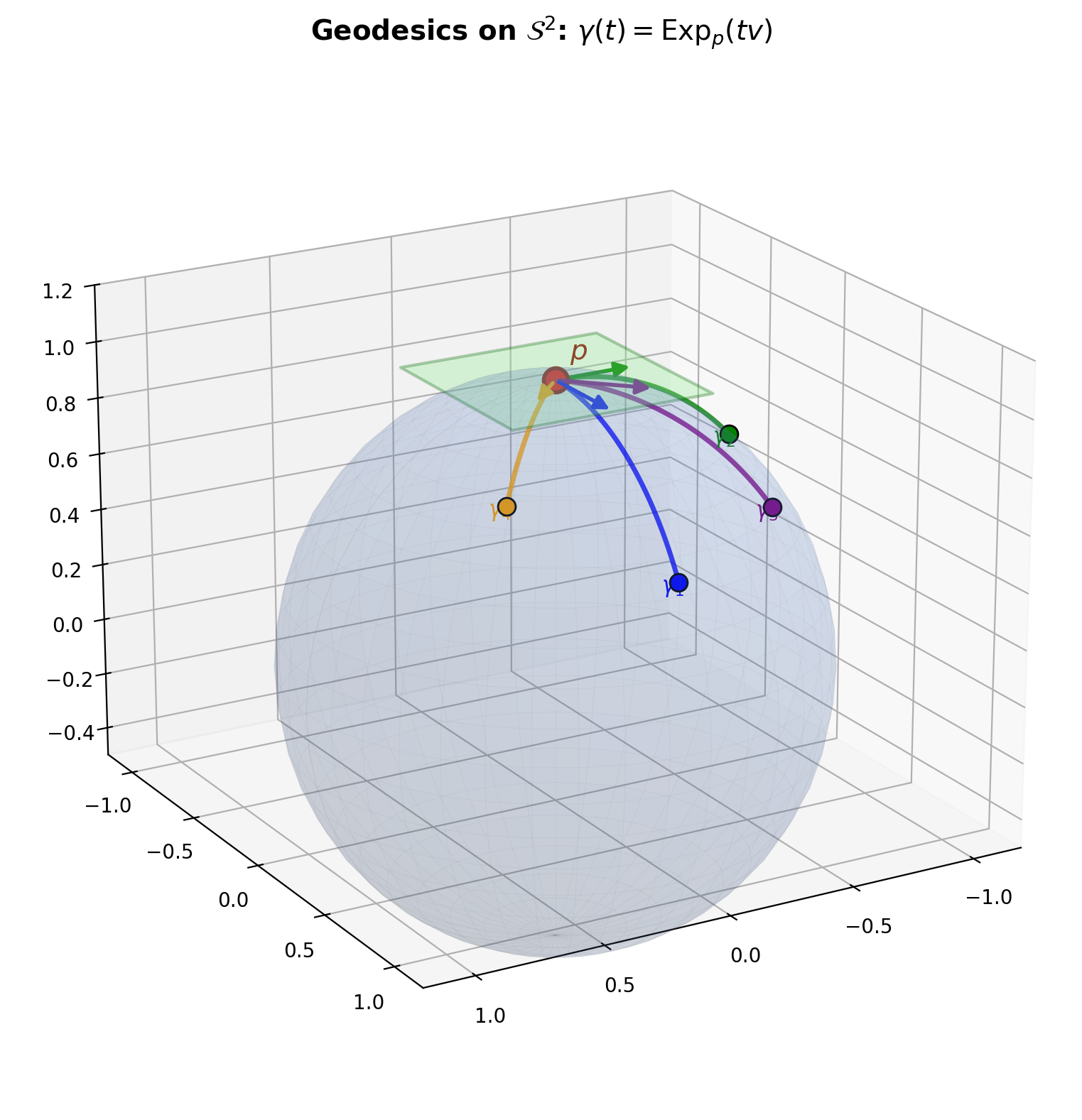}
        \caption{Geodesics on $\mathbb{S}^2$. From point $p$, geodesics $\gamma_i(t) = \mathrm{Exp}_p(tv_i)$ follow great circles determined by tangent vectors $v_i \in T_p\mathcal{M}$.}
        \label{fig:geodesics}
    \end{subfigure}
    \caption{Tangent space and geodesics on the sphere $\mathbb{S}^2$. (Left) The tangent space $T_p\mathcal{M}$ at point $p$. (Right) Geodesics emanating from $p$ induced by tangent vectors via the exponential map.}
    \label{fig:tangent_and_geodesics}
\end{figure}

\begin{figure}[h]
    \centering
    \includegraphics[width=\textwidth]{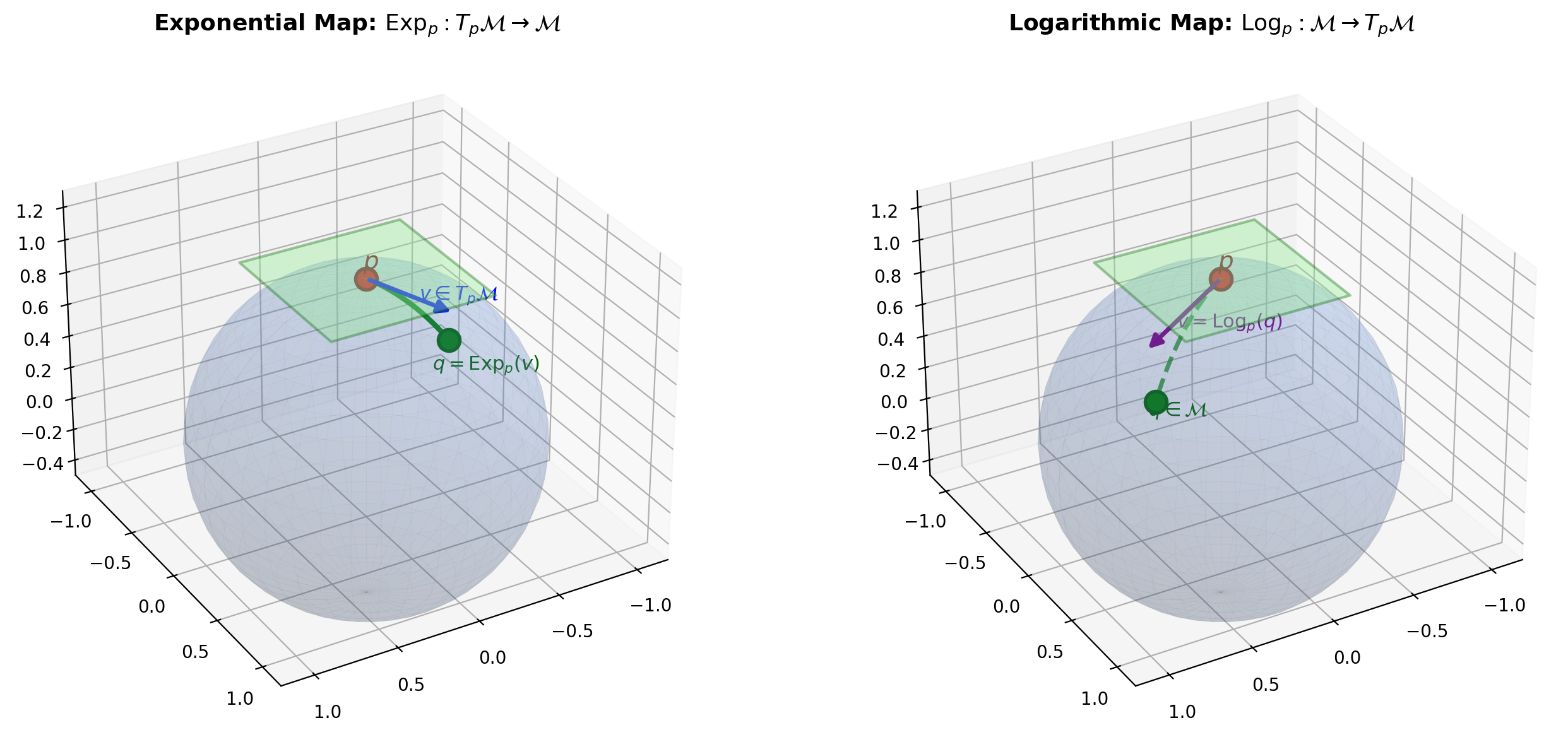}
    \caption{Exponential and logarithmic maps on $\mathbb{S}^2$. Left: $\mathrm{Exp}_p(v)$ maps tangent vector $v$ to point $q$ via the geodesic. Right: $\mathrm{Log}_p(q)$ returns the tangent vector whose exponential gives $q$.}
    \label{fig:exp_log_maps}
\end{figure}



\section{Optimization on Riemannian manifolds}
Optimization on manifolds arises naturally in problems with geometric constraints,
such as orthogonality, low-rank structure, and positive definiteness.
This section presents the basic concepts required for first-order optimization
on Riemannian manifolds, with particular emphasis on the definition and computation
of the Riemannian gradient.
Many constrained optimization problems can be formulated as unconstrained problems
over smooth manifolds. Classical Euclidean optimization techniques are not directly
applicable in this setting due to the lack of a global vector space structure.
Riemannian optimization addresses this issue by exploiting the differential-geometric
structure of the constraint set.\\

\nd Optimization on manifolds concerns problems of the form
\begin{equation}
\label{eq:manifold-optimization}
\min_{x \in \mathcal{M}} f(x),
\end{equation}
where $\mathcal{M}$ is a smooth manifold and $f : \mathcal{M} \to \mathbb{R}$ is a
smooth objective function. Such problems arise naturally when constraints define
a nonlinear geometric structure, as in orthogonality constraints, fixed-rank
matrix factorizations, and positive definite matrices.
\\
Unlike Euclidean spaces, manifolds generally lack a global vector space structure.
As a consequence, standard optimization methods based on linear updates are not
directly applicable. Instead, optimization algorithms must exploit the local
geometry of the manifold through its tangent spaces and a suitable Riemannian
metric.


%

\nd A Riemannian manifold $(\mathcal{M}, g)$ is a smooth manifold endowed with a
 Riemannian metric $g$, which assigns to each $x \in \mathcal{M}$ an inner product
 \[
 \langle \cdot, \cdot \rangle_x : T_x \mathcal{M} \times T_x \mathcal{M} \to \mathbb{R},
 \]
 varying smoothly with $x$.
The metric induces notions of length, angle, and orthogonality on the manifold.

\medskip
\begin{definition}[Differential of a Function]
Let $f : \mathcal{M} \to \mathbb{R}$ be a smooth function.
The differential of $f$ at $x \in \mathcal{M}$ is the linear map
\[
\mathrm{d}f(x) : T_x \mathcal{M} \to \mathbb{R},
\]
defined by the directional derivative of $f$ along tangent vectors.
\end{definition}

\nd 
The \emph{Riemannian gradient} of $f$ at $x$, denoted by $\operatorname{grad} f(x)$,
is the unique vector in $T_x \mathcal{M}$ satisfying
\begin{equation}
\label{eq:riemannian-gradient}
\langle \operatorname{grad} f(x), Z \rangle_x
=
\mathrm{d}f(x)[Z],
\quad \forall Z \in T_x \mathcal{M}.
\end{equation}
This definition generalizes the notion of the Euclidean gradient and characterizes
the direction of steepest ascent on the manifold with respect to the Riemannian metric.\\

\nd Suppose $\mathcal{M}$ is an embedded submanifold of $\mathbb{R}^n$ and the Riemannian
metric is induced by the Euclidean inner product.
Let $\tilde{f} : \mathbb{R}^n \to \mathbb{R}$ be a smooth extension of $f$.
Then the Riemannian gradient is given by
\[
\operatorname{grad} f(x)
=
\Pi_x \big( \nabla \tilde{f}(x) \big),
\]
where $\Pi_x$ denotes the orthogonal projection from $\mathbb{R}^n$ onto
$T_x \mathcal{M}$.
\\

\nd {\bf Example: Unit sphere}
Consider the unit sphere
\[
\mathbb{S}^{n-1} = \{ x \in \mathbb{R}^n : \|x\| = 1 \}.
\]
The tangent space at $x \in \mathbb{S}^{n-1}$ is
\[
T_x \mathbb{S}^{n-1} = \{ Z \in \mathbb{R}^n : x^\top Z = 0 \}.
\]
The orthogonal projection onto $T_x \mathbb{S}^{n-1}$ is
\[
\Pi_x(v) = v - (x^\top v)x,
\]
which yields the Riemannian gradient
\[
\operatorname{grad} f(x)
=
\nabla f(x) - (x^\top \nabla f(x))x.
\]
\emph{Riemannian Gradient Descent:} 
A basic first-order method on manifolds is Riemannian gradient descent, defined by
\[
x_{k+1} = \operatorname{Retr}_{x_k}
\left( -\alpha_k \operatorname{grad} f(x_k) \right),
\]
where $\alpha_k > 0$ is a step size and $\operatorname{Retr}_{x_k}$ is a retraction
mapping tangent vectors back onto the manifold.
\\
The Riemannian gradient provides a natural extension of the Euclidean gradient to
manifold-constrained optimization problems. It forms the foundation for a wide class
of first- and second-order algorithms in Riemannian optimization.

\nd We state the standard assumptions under which first-order optimization methods
on Riemannian manifolds are guaranteed to converge.
\\

\nd Throughout this subsection, we make the following assumptions.

\begin{enumerate}
\item {Smoothness}:
The objective function $f : \mathcal{M} \to \mathbb{R}$
is smooth (infinitely differentiable) on $\mathcal{M}$.
\item ({Lower boundedness})
The function $f$ is bounded below on $\mathcal{M}$, i.e., there exists
$f^\star \in \mathbb{R}$ such that
\[
f(x) \geq f^\star, \quad \forall x \in \mathcal{M}.
\]
\item  
The Riemannian gradient of $f$ is $L$-Lipschitz continuous, meaning that
for all $x \in \mathcal{M}$ and all $Z \in T_x \mathcal{M}$,
\[
\left\|
\operatorname{grad} f(y) - \mathcal{P}_{x \to y}\operatorname{grad} f(x)
\right\|
\leq L \, \mathrm{dist}(x,y),
\]
where $y = \operatorname{Exp}_x(Z)$, $\mathcal{P}_{x \to y}$ denotes
parallel transport along the geodesic from $x$ to $y$, and
$\mathrm{dist}(\cdot,\cdot)$ is the Riemannian distance.
\item 
The retraction $\operatorname{Retr}$ is first-order accurate, i.e.,
\[
\operatorname{Retr}_x(0) = x, \qquad
\mathrm{D}\operatorname{Retr}_x(0) = \mathrm{Id}.
\]
\end{enumerate}

\begin{theorem}
\label{thm:rgd-convergence}
Suppose Assumptions \textbf{1}--\textbf{4} hold and the step sizes satisfy
$0 < \alpha_k \leq \alpha < 2/L$.
Then the sequence $\{x_k\}$ generated by Riemannian gradient descent satisfies
\[
\lim_{k \to \infty} \| \operatorname{grad} f(x_k) \| = 0.
\]
In particular, every accumulation point of $\{x_k\}$ is a first-order
critical point of $f$.
\end{theorem}

\begin{theorem}
\label{thm:rgd-rate}
Let Assumptions \textbf{1}--\textbf{4} hold and choose a constant step size
$\alpha = 1/L$.
Then after $K$ iterations, the iterates satisfy
\[
\min_{0 \leq k < K}
\| \operatorname{grad} f(x_k) \|^2
\leq
\frac{2L \big( f(x_0) - f^\star \big)}{K}.
\]
\end{theorem}

\nd The above results show that Riemannian gradient descent enjoys convergence
properties analogous to its Euclidean counterpart.
Under standard smoothness assumptions, the algorithm converges to first-order
stationary points at a rate of $\mathcal{O}(1/K)$.
Stronger guarantees, such as linear convergence, can be obtained under additional
assumptions including geodesic convexity or the Polyak--Łojasiewicz condition
on manifolds.

	\section{Principal Geodesic Analysis (PGA)}

Principal Geodesic Analysis (PGA) is the natural generalization of Principal Component Analysis (PCA) to data that lies on a Riemannian manifold. While PCA identifies directions of maximal variance in Euclidean space, PGA identifies directions of maximal variation along geodesics on curved spaces, respecting the manifold's intrinsic geometry.

Let $\mathcal{M}$ be a Riemannian manifold, and let $x_1, \dots, x_N \in \mathcal{M}$ be a set of data points. Since these points lie on a curved space, standard linear projections are not directly applicable. PGA overcomes this by first selecting a reference point on the manifold and then performing PCA in the tangent space at that point.
\\
The reference point is the \textbf{Fr\'echet (Karcher) mean} $x_0 \in \mathcal{M}$, which minimizes the sum of squared geodesic distances to all data points:
\[
x_0 = \arg\min_{x \in \mathcal{M}} \sum_{i=1}^{N} d_\mathcal{M}(x, x_i)^2,
\]
where $d_\mathcal{M}(x, y)$ denotes the geodesic distance between $x$ and $y$ on the manifold. The Fr\'echet mean provides a central reference around which variations are measured.
\\
Once the mean is computed, each point is mapped to the tangent space at $x_0$ using the \textbf{logarithmic map}:
\[
z_i = \operatorname{Log}_{x_0}(x_i) \in T_{x_0}\mathcal{M}, \quad i = 1, \dots, N.
\]
The tangent space $T_{x_0}\mathcal{M}$ is a linear space, allowing standard linear algebra operations such as covariance computation and eigen decomposition. In the tangent space, the covariance matrix is defined as
\[
\Sigma = \frac{1}{N} \sum_{i=1}^{N} z_i z_i^\top.
\]
Eigen-decomposition of $\Sigma$ yields
\[
\Sigma u_j = \lambda_j u_j, \quad j = 1, \dots, d,
\]
where the eigenvectors $u_j$ represent the \textbf{principal geodesics} (directions of maximal variation on the manifold) and the eigenvalues $\lambda_j$ quantify the variance along each geodesic.
\\
For dimensionality reduction, the top $k \ll d$ eigenvectors are selected, and each tangent vector is projected onto these directions:
\[
\tilde{z}_i =
\begin{bmatrix}
u_1^\top z_i \\
\vdots \\
u_k^\top z_i
\end{bmatrix} \in \mathbb{R}^k, \quad i = 1, \dots, N.
\]
If desired, these low-dimensional representations can be mapped back to the manifold via the exponential map:
\[
\tilde{x}_i = \operatorname{Exp}_{x_0} \Bigg( \sum_{j=1}^k (u_j^\top z_i) u_j \Bigg) \in \mathcal{M}.
\]

In the special case of data lying on the unit hypersphere $\mathbb{S}^{d-1}$, such as normalized token embeddings. the geodesic distance between points $x, y \in \mathbb{S}^{d-1}$ is
\[
d(x, y) = \arccos(x^\top y),
\]
and the logarithmic map admits the closed form
\[
\operatorname{Log}_{x_0}(x) = \frac{\theta}{\sin \theta} (x - \cos\theta \, x_0), \quad \theta = \arccos(x_0^\top x).
\]
The tangent-space covariance can then be computed, and the principal geodesics obtained by standard eigen-decomposition. To reconstruct the geodesic on the sphere corresponding to a principal direction $u_k$, the exponential map is used:
\[
\gamma_k(t) = \operatorname{Exp}_{x_0}(t \, u_k) = \cos(t \|u_k\|) x_0 + \sin(t \|u_k\|) \frac{u_k}{\|u_k\|}.
\]
PGA is particularly useful in settings, where token embeddings naturally lie on a hypersphere. By identifying dominant geodesic directions, redundant tokens can be detected and compressed, while preserving the semantic structure of the data. 
\\
Moreover, PGA can be extended to other Riemannian manifolds, such as the manifold of symmetric positive definite (SPD) matrices $\mathcal{S}_{++}^d$. For SPD matrices $C_i$, the tangent-space mapping is
\[
Z_i = C_0^{1/2}\operatorname{log}(C_0^{-1/2} C_i C_0^{-1/2})C_0^{1/2},
\]
the covariance in the tangent space is
\[
\Sigma = \frac{1}{N} \sum_{i=1}^N Z_i Z_i^\top,
\]
and dimensionality-reduced representations can be mapped back to the manifold via
\[
\tilde{C}_i = C_0^{1/2}  \operatorname{exp}\Big(C_0^{-1/2}(\sum_j (u_j^\top Z_i) u_j) C_0^{-1/2}\Big) C_0^{1/2}.
\]
This enables geometry-aware dimensionality reduction while respecting the Riemannian structure of the SPD manifold.
\\
Let 
\[
Z = [z_1, \dots, z_N] \in \mathbb{R}^{d \times N}
\] 
be the tangent vectors at \(x_0 \in \mathcal{M}\), where 
\[
z_i = \operatorname{Log}_{x_0}(x_i), \quad i = 1, \dots, N.
\] 
The covariance matrix in the tangent space is
\[
\Sigma = \frac{1}{N} \sum_{i=1}^{N} z_i z_i^\top = \frac{1}{N} Z Z^\top.
\]
PGA seeks the top \(k\) principal geodesic directions 
\[
U = [u_1, \dots, u_k] \in \mathbb{R}^{d \times k},
\] 
where the columns are orthonormal (\(U^\top U = I_k\)), that maximize the variance of the projected data.
\\
This is equivalent to the trace maximization problem:
\begin{equation}
\begin{aligned}
\max_{U \in \mathbb{R}^{d \times k}} & \quad \mathrm{tr}(U^\top \Sigma U) \\
\text{subject to} & \quad U^\top U = I_k.
\end{aligned}
\end{equation}
\\
Note that 
\[
\mathrm{tr}(U^\top \Sigma U) = \sum_{j=1}^k u_j^\top \Sigma u_j
\] 
measures the total variance along the selected directions. The orthonormality constraint ensures that the directions are independent in the tangent space.
\\
The solution is obtained by eigen-decomposition of \(\Sigma\), selecting the top \(k\) eigenvectors corresponding to the largest eigenvalues.

	

\begin{algorithm}[t]
\caption{Riemannian PCA (Principal Geodesic Analysis)}
\label{alg:rpca}
\textbf{Input:}
Data points $\{x_i\}_{i=1}^N \subset \mathcal{M}$ and target dimension $d$.\\
\textbf{Output:}
Principal directions $\{v_1,\dots,v_d\}$ and reduced coordinates
$\{z_i\}_{i=1}^N$.
\medskip
\begin{enumerate}
\item {Computation of the Fréchet mean.}
Initialize $\bar{x}_0 \in \mathcal{M}$.
For $k=0,\dots,K_{\mathrm{mean}}$, update
\[
\bar{x}_{k+1}
=
\operatorname{Exp}_{\bar{x}_k}
\left(
\frac{1}{N}
\sum_{i=1}^N
\mathcal{T}_{\bar{x}_k}(x_i)
\right),
\]
where $\mathcal{T}_{\bar{x}_k}(\cdot)$ denotes a smooth tangent lifting map.
Set $\bar{x}=\bar{x}_{K_{\mathrm{mean}}}$.
\item {Tangent space representation.}
For each data point $x_i$, compute its tangent representation
\[
Z_i = \mathcal{T}_{\bar{x}}(x_i) \in T_{\bar{x}}\mathcal{M}.
\]
\item {Covariance operator.}
Form the empirical covariance operator
\[
C = \frac{1}{N} \sum_{i=1}^N Z_i \otimes Z_i .
\]
\item {Spectral decomposition.}
Compute the $d$ dominant eigenvectors $\{v_1,\dots,v_d\}$ of $C$.
\item {Low-dimensional embedding.}
For each $i=1,\dots,N$, define the reduced coordinates
\[
z_i =
\big(
\langle Z_i,v_1\rangle,
\dots,
\langle Z_i,v_d\rangle
\big)^\top .
\]
\end{enumerate}
\end{algorithm}

\section{Riemannian Robust Principal Component Analysis (RRPCA)}
Robust Principal Component Analysis (RPCA) aims to decompose data into a low-rank component and a sparse corruption component. 
When data lie on a Riemannian manifold rather than in Euclidean space, classical RPCA formulations are no longer directly applicable.
Riemannian Robust PCA (RRPCA) extends RPCA to manifold-valued data by respecting the intrinsic geometry through tangent-space representations and Riemannian optimization.
\\
Let $\{x_i\}_{i=1}^N \subset \mathcal{M}$ be data points lying on a Riemannian manifold $(\mathcal{M}, g)$.
The goal of RRPCA is to separate the data into a low-dimensional manifold-consistent structure and sparse outliers.

\paragraph{Reference Point and Tangent Space Mapping.}
A reference point $x_0 \in \mathcal{M}$ is selected, typically as the Fr\'echet mean:
\[
x_0 = \arg\min_{x \in \mathcal{M}} \sum_{i=1}^N d_{\mathcal{M}}^2(x, x_i),
\]
where $d_{\mathcal{M}}(\cdot,\cdot)$ denotes the geodesic distance.
Each data point is mapped to the tangent space at $x_0$ via the logarithmic map:
\[
z_i = \operatorname{Log}_{x_0}(x_i) \in T_{x_0}\mathcal{M}.
\]
The tangent space $T_{x_0}\mathcal{M}$ is a vector space, enabling linear decomposition.
\paragraph{Riemannian robust RPCA Formulation.}
Let $Z = [z_1, \dots, z_N] \in \mathbb{R}^{d \times N}$ denote the tangent-space data matrix.
RRPCA seeks a decomposition
\[
Z = L + S,
\]
where $L$ is a low-rank matrix capturing the intrinsic structure, and $S$ is a sparse matrix modeling outliers or gross corruptions.
\\
The Riemannian robust PCA optimization problem is formulated as
\[
\min_{L,S} \quad \|L\|_* + \lambda \|S\|_1
\quad \text{subject to} \quad
Z = L + S,
\]
where $\|\cdot\|_*$ denotes the nuclear norm and $\|\cdot\|_1$ the element-wise $\ell_1$ norm.
The parameter $\lambda > 0$ balances low-rankness and sparsity.
\\
The above problem can be efficiently solved using an Alternating Direction Method of Multipliers (ADMM) framework.
The low-rank update involves Singular Value Thresholding (SVT) in the tangent space:
\[
L^{(k+1)} = \mathcal{T}_{\tau}(Z - S^{(k)} + Y^{(k)}),
\]
where $\mathcal{T}_{\tau}(\cdot)$ denotes the soft-thresholding of singular values.
This step is often referred to as \emph{Riemannian Singular Value Thresholding (RSVT)} when coupled with manifold-aware mappings.
\\
After obtaining the low-rank tangent representation $\hat{L} = [\hat{\ell}_1, \dots, \hat{\ell}_N]$, the cleaned manifold-valued data are recovered via the exponential map:
\[
\hat{x}_i = \operatorname{Exp}_{x_0}(\hat{\ell}_i), \quad i = 1, \dots, N.
\]
\\
RRPCA identifies a low-dimensional geodesic subspace of the manifold that best represents the data while discarding sparse deviations.
Unlike Euclidean RPCA, RRPCA preserves intrinsic distances and curvature, ensuring that the recovered structure remains consistent with the manifold geometry.

\nd \emph{Special Cases.}
\begin{itemize}
\item {Hypersphere $\mathbb{S}^{d-1}$:} Tangent vectors are orthogonal to the mean direction, and logarithmic and exponential maps have closed forms.
\item {SPD Manifold $\mathcal{S}_{++}^d$:} The tangent space consists of symmetric matrices, and logarithmic/exponential maps are defined via matrix logarithms and exponentials.
\item {Grassmann Manifold.} RRPCA captures dominant subspaces while being invariant to basis rotations.
\end{itemize}

\begin{remark}
RRPCA generalizes classical RPCA to curved spaces and is particularly well suited for applications involving covariance matrices, directional data, and subspace-valued observations.
In the Euclidean case, where $\mathcal{M} = \mathbb{R}^d$, the formulation reduces to standard RPCA.
\end{remark}

\section{ONPP on Riemannian manifolds}

The classical Orthogonal Neighborhood Preserving Projections (ONPP) is a linear dimensionality
reduction method that seeks an orthogonal projection preserving local neighborhood
relationships. Given data points $\{x_i\}_{i=1}^N \subset \mathbb{R}^n$, ONPP computes
a weight matrix $W$ by reconstructing each point from its neighbors and then finds
an orthogonal projection matrix $U \in \mathbb{R}^{n \times d}$ solving
\[
\min_{U^\top U = I_d} \sum_{i=1}^N
\left\| U^\top x_i - \sum_{j=1}^N W_{ij} U^\top x_j \right\|^2.
\]
This leads to an eigenvalue problem involving the data covariance weighted by
local reconstruction coefficients.
 When data lie on a Riemannian manifold $\mathcal{M}$, Euclidean subtraction
$x_i - x_j$ is no longer well-defined. To extend ONPP to manifolds, we exploit
the local linear structure of tangent spaces.
\\
Let $x_i \in \mathcal{M}$ be data points. For each $x_i$, define its neighborhood
$\mathcal{N}(i)$ and map neighboring points to the tangent space $T_{x_i}\mathcal{M}$
using the logarithm map:
\[
Z_{ij} = \operatorname{Log}_{x_i}(x_j), \quad j \in \mathcal{N}(i).
\]

\nd The reconstruction weights $W_{ij}$ are obtained by solving, for each $i$,
\[
\min_{\{W_{ij}\}}
\left\|
\sum_{j \in \mathcal{N}(i)} W_{ij} \, Z_{ij}
\right\|^2
\quad \text{s.t.} \quad
\sum_{j \in \mathcal{N}(i)} W_{ij} = 1.
\]
This formulation preserves the local geometry of the manifold up to first order,
since the logarithm map provides a locally isometric representation.

\nd \emph{Projection on the Stiefel Manifold}
Let $X_i \in \mathbb{R}^n$ denote an embedding or representation of $x_i$
(e.g., via coordinates or ambient embedding).
The Riemannian ONPP objective can be written as
\[
\min_{U \in \mathrm{St}(n,d)}
\sum_{i=1}^N
\left\|
U^\top X_i - \sum_{j=1}^N W_{ij} U^\top X_j
\right\|^2,
\]
where
\[
\mathrm{St}(n,d) = \{ U \in \mathbb{R}^{n \times d} : U^\top U = I_d \}
\]
is the Stiefel manifold.
\\
This objective can be expressed compactly as
\[
\min_{U \in \mathrm{St}(n,d)} \;
\operatorname{tr}\left( U^\top X M X^\top U \right),
\]
where $M = (I - W)^\top (I - W)$.

\nd The optimization problem is solved using Riemannian optimization techniques on
the Stiefel manifold. The Riemannian gradient of the objective function
\[
f(U) = \operatorname{tr}\left( U^\top X M X^\top U \right)
\]
is given by
\[
\operatorname{grad} f(U)
=
\nabla f(U) - U \operatorname{sym}(U^\top \nabla f(U)),
\]
where
\[
\nabla f(U) = 2 X M X^\top U,
\quad
\operatorname{sym}(A) = \tfrac{1}{2}(A + A^\top).
\]
The update rule is
\[
U_{k+1} = \operatorname{Retr}_{U_k}
\left( -\alpha_k \operatorname{grad} f(U_k) \right),
\]
where $\operatorname{Retr}$ is a retraction on the Stiefel manifold, such as the
QR-based retraction.
\\

\nd Under standard smoothness assumptions on the objective function and a suitable
choice of step sizes, Riemannian gradient-based ONPP converges to a first-order
stationary point on the Stiefel manifold. These guarantees follow directly from
classical results in Riemannian optimization.
\\
By combining tangent-space neighborhood preservation with orthogonality-constrained
optimization, the Riemannian ONPP method respects both the intrinsic geometry of
manifold-valued data and the global structure imposed by orthogonal projections.
This makes it particularly suitable for dimensionality reduction of data lying
on nonlinear manifolds.

\subsection{ONPP on the SPD manifold}

Let $\mathcal{S}_{++}^n$ denote the manifold of $n \times n$ symmetric positive
definite (SPD) matrices. Equipped with the affine-invariant Riemannian metric,
$\mathcal{S}_{++}^n$ forms a smooth Riemannian manifold.

\nd {\bf Geometry of $\mathcal{S}_{++}^n$}
For $X \in \mathcal{S}_{++}^n$, the tangent space is
\[
T_X \mathcal{S}_{++}^n = \{ Z \in \mathbb{R}^{n \times n} : Z = Z^\top \}.
\]
The affine-invariant Riemannian metric is given by
\[
\langle Z, \eta \rangle_X
=
\operatorname{tr}\left( X^{-1} Z X^{-1} \eta \right).
\]
The logarithm map at $X$ is
\[
\operatorname{Log}_X(Y)
=
X^{1/2} \operatorname{log}\left( X^{-1/2} Y X^{-1/2} \right) X^{1/2}.
\]

\nd {Neighborhood Reconstruction}
Given SPD data points $\{X_i\}_{i=1}^N \subset \mathcal{S}_{++}^n$, neighborhood
weights are computed by solving
\[
\min_{\{W_{ij}\}}
\left\|
\sum_{j \in \mathcal{N}(i)}
W_{ij} \operatorname{Log}_{X_i}(X_j)
\right\|_F^2,
\quad
\text{s.t. }
\sum_{j \in \mathcal{N}(i)} W_{ij} = 1.
\]

\nd {Projection Learning}
Each SPD matrix $X_i$ is vectorized in a common tangent space (e.g., at the
Karcher mean $\bar{X}$) via
\[
Z_i = \operatorname{Log}_{\bar{X}}(X_i).
\]
Let $Z = [\mathrm{vec}(Z_1), \dots, \mathrm{vec}(Z_N)]$.
The ONPP projection matrix $U \in \mathbb{R}^{n^2 \times d}$ is obtained by solving
\[
\min_{U \in \mathrm{St}(n^2,d)}
\operatorname{tr}\left( U^\top Z M Z^\top U \right),
\]
where $M = (I - W)^\top (I - W)$.

\subsection{ONPP on the Grassmann manifold}

Let $\mathrm{Gr}(p,n)$ denote the Grassmann manifold of $p$-dimensional subspaces
of $\mathbb{R}^n$.

\nd {Geometry of $\mathrm{Gr}(p,n)$}
A point on $\mathrm{Gr}(p,n)$ is represented by an orthonormal matrix
$X \in \mathbb{R}^{n \times p}$ such that $X^\top X = I_p$.
The tangent space at $X$ is
\[
T_X \mathrm{Gr}(p,n)
=
\{ Z \in \mathbb{R}^{n \times p} : X^\top Z = 0 \}.
\]
The logarithm map between two points $X, Y \in \mathrm{Gr}(p,n)$ is given by
\[
\operatorname{Log}_X(Y) = U \arctan(\Sigma) V^\top,
\]
where $U \Sigma V^\top$ is obtained from the compact SVD of
$(I - XX^\top)Y (X^\top Y)^{-1}$.

\nd {Neighborhood Reconstruction}
For each $X_i \in \mathrm{Gr}(p,n)$, neighbors are mapped to $T_{X_i}\mathrm{Gr}(p,n)$:
\[
Z_{ij} = \operatorname{Log}_{X_i}(X_j).
\]
The reconstruction weights are computed by
\[
\min_{\{W_{ij}\}}
\left\|
\sum_{j \in \mathcal{N}(i)} W_{ij} Z_{ij}
\right\|_F^2,
\quad
\text{s.t. }
\sum_{j \in \mathcal{N}(i)} W_{ij} = 1.
\]

\nd {Projection Learning}
Let $X_i$ be represented in an ambient space (or concatenated tangent
representation). The ONPP objective becomes
\[
\min_{U \in \mathrm{St}(n,d)}
\operatorname{tr}\left( U^\top X M X^\top U \right),
\]
which is solved via Riemannian optimization on the Stiefel manifold.

\begin{algorithm}[h]
\caption{Riemannian Orthogonal Neighborhood Preserving Projection (R-ONPP)}
\label{alg:ronpp}

\textbf{Input:}
Manifold-valued data $\{x_i\}_{i=1}^N \subset \mathcal{M}$,
neighborhood size $k$, and target dimension $d$.
\\
\textbf{Output:}
Projection matrix $U \in \mathbb{R}^{n \times d}$.
\medskip
\begin{enumerate}
\item 
For each data point $x_i$, construct a neighborhood
$\mathcal{N}(i)$ using a distance on $\mathcal{M}$.
\item 
For each $i=1,\dots,N$ and $j \in \mathcal{N}(i)$, compute a local tangent
representation
\[
Z_{ij} = \mathcal{T}_{x_i}(x_j),
\]
where $\mathcal{T}_{x_i}(\cdot)$ denotes a smooth tangent lifting or retraction-based inverse.
\item 
For each $i=1,\dots,N$, determine weights $\{W_{ij}\}_{j \in \mathcal{N}(i)}$
by solving
\[
\min_{\{W_{ij}\}}
\left\|
\sum_{j \in \mathcal{N}(i)} W_{ij} Z_{ij}
\right\|^2,
\qquad
\text{subject to }
\sum_{j \in \mathcal{N}(i)} W_{ij} = 1 .
\]
\item 
Assemble the global weight matrix $W$ and form
\[
M = (I - W)^\top (I - W).
\]
\item 
Embed all data points into a common vector space, for example the tangent
space at the Fréchet mean, yielding vectors $\{Z_i\}_{i=1}^N \subset \mathbb{R}^n$.
\item 
Initialize the projection matrix $U_0 \in \mathrm{St}(n,d)$ on the Stiefel manifold.
\item {Riemannian optimization.}
For $k = 0,\dots,K$, iterate
\[
U_{k+1}
=
\operatorname{Retr}_{U_k}
\big(
-\alpha_k \operatorname{grad} f(U_k)
\big),
\]
where the objective function is: 
$ 
f(U) = \operatorname{trace}(U^\top Z M Z^\top U),
$ 
\end{enumerate}
\end{algorithm}

\newpage

\section{Riemannian Laplacian Eigenmaps}
Laplacian Eigenmaps (LE) is a spectral embedding technique that seeks to preserve the local geometry of the data. For data points $\{x_i\}_{i=1}^N$ lying on a
Riemannian manifold $\mathcal{M}$, the key idea is to build a neighborhood graph
that captures the manifold structure and compute a low-dimensional representation
by solving a generalized eigenvalue problem associated with the graph Laplacian.


\nd Let $\mathcal{N}(i)$ denote the $k$ nearest neighbors of $x_i$ measured by
the geodesic distance on $\mathcal{M}$:
\[
d_\mathcal{M}(x_i, x_j) = \mathrm{dist}(x_i, x_j)
=
\inf_\gamma \int_0^1 \sqrt{ g_{\gamma(t)}(\dot{\gamma}(t), \dot{\gamma}(t)) } \, dt,
\]
where $\gamma: [0,1] \to \mathcal{M}$ is a smooth curve connecting $x_i$ to $x_j$
and $g$ is the Riemannian metric.  

This graph encodes local adjacency and ensures that only nearby points influence
the embedding, avoiding distortions caused by the curvature of the manifold.


\nd The adjacency matrix $W$ assigns weights to edges based on local distances:

\begin{equation}
\label{eq:manifold-weights}
W_{ij} =
\begin{cases}
\operatorname{exp}\left(-\frac{d_\mathcal{M}(x_i,x_j)^2}{t}\right), & x_j \in \mathcal{N}(i) \\
0, & \text{otherwise},
\end{cases}
\end{equation}

where $t>0$ is a heat kernel scaling parameter.  
Alternative formulations include binary weights $W_{ij}=1$ for neighbors.


\nd Define the degree matrix $D$ as
\[
D_{ii} = \sum_j W_{ij}.
\]
The (unnormalized) graph Laplacian is
\[
L = D - W.
\]

LE seeks embeddings $\{y_i\}_{i=1}^N \subset \mathbb{R}^d$ that minimize the cost
function
\begin{equation}
\label{eq:LE-objective}
\sum_{i,j} W_{ij} \| y_i - y_j \|^2
= \mathrm{tr}(Y^\top L Y),
\end{equation}
subject to a scale-fixing constraint $Y^\top D Y = I_d$, where
$Y = [y_1^\top; \dots; y_N^\top] \in \mathbb{R}^{N \times d}$.  

\nd The solution corresponds to the $d$ eigenvectors associated with the smallest
nonzero eigenvalues of the generalized eigenvalue problem:
\[
L Y = D Y \Lambda.
\]


\nd While the above formulation is formally similar to Euclidean LE, the manifold
setting requires careful treatment:

\begin{enumerate}
    \item {Geodesic distances:} Use intrinsic distances $d_\mathcal{M}(x_i,x_j)$
    rather than Euclidean distances in the ambient space.
    
    \item {Tangent space approximation:} For manifolds with computationally
    expensive geodesics (e.g., Grassmann or SPD manifolds), neighborhoods can be
    linearized via the logarithm map:
    \[
    Z_{ij} = \operatorname{Log}_{x_i}(x_j) \in T_{x_i}\mathcal{M},
    \]
    and local distances are approximated as $\|Z_{ij}\|$.
    
    \item {Weight symmetrization:} In general, $d_\mathcal{M}(x_i,x_j) \neq d_\mathcal{M}(x_j,x_i)$
    numerically. Symmetrize $W$ via $W \leftarrow (W + W^\top)/2$.
    
    \item {Curvature effects:} LE assumes local neighborhoods are approximately
    Euclidean. For manifolds with high curvature, neighborhood size $k$ should
    remain small to preserve the local linear structure.
\end{enumerate}


\nd For new points $x \in \mathcal{M}$, LE does not automatically provide an embedding.
One common strategy is the Nystrom extension:

\begin{equation}
\label{eq:nystrom}
y = \frac{1}{\sum_i W_i} \sum_i W_i y_i,
\quad W_i = \operatorname{exp}\left(-\frac{d_\mathcal{M}(x, x_i)^2}{t}\right),
\end{equation}
where $y_i$ are the learned embeddings of neighbors $x_i \in \mathcal{N}(x)$.

\begin{remark}
\begin{itemize}
    \item For SPD manifolds, geodesic distances are computed via the affine-invariant metric.
    \item For Grassmann manifolds, geodesic distances can be computed using principal angles.
    \item Laplacian Eigenmaps can be combined with tangent-space linearizations for
    hybrid methods that improve scalability for large datasets.
    \item Normalized Laplacians $L_\text{sym} = I - D^{-1/2} W D^{-1/2}$ can
    also be used to improve embedding stability.
\end{itemize}
\end{remark}

\begin{algorithm}[h]
\caption{Riemannian Laplacian Eigenmaps}
\label{alg:le-detailed}
\textbf{Input:} Manifold data $\{x_i\}_{i=1}^N \subset \mathcal{M}$, 
target dimension $d$, neighborhood size $k$, kernel scale $t$.
\\
\textbf{Output:} Embeddings $\{y_i\}_{i=1}^N \subset \mathbb{R}^d$.
\begin{enumerate}
    \item Construct the neighborhood $\mathcal{N}(i)$ for each $x_i$ based on the manifold distance $d_{\mathcal{M}}$.
    \item Compute the weight matrix $W$ using Eq.~\eqref{eq:manifold-weights}.
    \item Symmetrize the weight matrix as $W \leftarrow (W + W^\top)/2$.
    \item Compute the degree matrix $D$ and the graph Laplacian $L = D - W$.
    \item Solve the generalized eigenvalue problem $L Y = D Y \Lambda$.
    \item Select the $d$ eigenvectors corresponding to the smallest nonzero eigenvalues.
    \item \emph{Optional:} For an out-of-sample point $x \in \mathcal{M}$, compute its embedding using Eq.~\eqref{eq:nystrom}.
\end{enumerate}
\end{algorithm}



\nd Riemannian Laplacian Eigenmaps generalize classical LE to manifold-valued data by
preserving local geometric structure in the embedding. It is particularly effective
when the manifold structure is non-Euclidean, as in SPD matrices, rotation groups,
or Grassmannians. This approach provides a nonlinear, geometry-aware embedding
that complements tangent-space methods such as Riemannian PCA.

\section{Linear discriminant analysis on Riemannian manifolds}
Linear Discriminant Analysis (LDA) is a classical supervised dimensionality reduction
method in Euclidean space that maximizes the between-class scatter while minimizing
the within-class scatter. For data lying on a Riemannian manifold, LDA can be
generalized to respect the geometry of the manifold, resulting in Riemannian LDA (RLDA).
\\

\nd Let $\{x_i, y_i\}_{i=1}^N \subset \mathcal{M} \times \{1,\dots,C\}$ be data points on a
Riemannian manifold $\mathcal{M}$ with class labels $y_i \in \{1,\dots,C\}$.  
The goal is to find a low-dimensional representation that maximizes class separability.\\

\nd \emph{Tangent space representation:}
Compute the Fréchet mean of the entire dataset:
\[
\bar{x} = \arg\min_{x \in \mathcal{M}} \sum_{i=1}^N \mathrm{dist}^2(x, x_i).
\]
Map all points to the tangent space at the mean:
\[
Z_i = \operatorname{Log}_{\bar{x}}(x_i) \in T_{\bar{x}}\mathcal{M}.
\]
Within each class $c$, define the class mean in the tangent space:
\[
\bar{Z}_c = \frac{1}{N_c} \sum_{i:y_i=c} Z_i,
\]
where $N_c$ is the number of points in class $c$.


\nd Define the between-class scatter:
\[
S_B = \sum_{c=1}^C N_c (\bar{Z}_c - \bar{Z})(\bar{Z}_c - \bar{Z})^\top,
\quad
\bar{Z} = \frac{1}{N} \sum_{i=1}^N Z_i.
\]

\nd Define the within-class scatter:
\[
S_W = \sum_{c=1}^C \sum_{i:y_i=c} (Z_i - \bar{Z}_c)(Z_i - \bar{Z}_c)^\top.
\]
The tangent-space LDA seeks a projection matrix $U \in \mathbb{R}^{\dim(T_{\bar{x}}\mathcal{M}) \times d}$ that maximizes the generalized Rayleigh quotient:
\[
\max_{U^\top U = I_d} \frac{\operatorname{tr}(U^\top S_B U)}{\operatorname{tr}(U^\top S_W U)}.
\]
The solution is given by the leading eigenvectors of
\[
S_W^{-1} S_B.
\]
After computing the projection $U$, the low-dimensional representation of $x_i$ is
\[
z_i = U^\top Z_i \in \mathbb{R}^d.
\]
Optionally, a reconstruction in the manifold can be obtained via the exponential map:
\[
\hat{x}_i = \operatorname{Exp}_{\bar{x}}(U z_i).
\]

\nd Some Special Cases:

\begin{itemize}
    \item \emph{SPD manifold:} Map SPD matrices $X_i \in \mathcal{S}_{++}^n$ to the tangent space at the Fréchet mean $\bar{X}$ using
    \[
    Z_i = \operatorname{Log}_{\bar{X}}(X_i) = \bar{X}^{1/2} \operatorname{log}(\bar{X}^{-1/2} X_i \bar{X}^{-1/2}) \bar{X}^{1/2}.
    \]
    Then compute $S_W$ and $S_B$ as above.
    
    \item \emph{Grassmann manifold:} For subspaces $X_i \in \mathrm{Gr}(p,n)$, map to the tangent space using the Grassmann logarithm map. Then compute scatter matrices in the tangent space.
\end{itemize}
\newpage 

\begin{algorithm}[h]
\caption{Riemannian Linear Discriminant Analysis}
\label{alg:rllda}
\textbf{Input:} Manifold data $\{x_i, y_i\}_{i=1}^N \subset \mathcal{M} \times \{1,\dots,C\}$, target dimension $d$.
\\
\textbf{Output:} Projection matrix $U \in \mathbb{R}^{\dim(T_{\bar{x}}\mathcal{M}) \times d}$ and low-dimensional embeddings $\{z_i\}$.
\begin{enumerate}
    \item Compute the Fréchet mean: 
    \( 
    \bar{x} = \displaystyle \arg\min_{x \in \mathcal{M}} \sum_{i=1}^N \mathrm{dist}^2(x, x_i).
    \)
    \item Map all data points to the tangent space at $\bar{x}$:
    \[
    Z_i = \operatorname{Log}_{\bar{x}}(x_i).
    \]
    \item For each class $c = 1,\dots,C$, compute the class mean in the tangent space:
   $ 
    \bar{Z}_c = \frac{1}{N_c} \sum_{i : y_i = c} Z_i.
   $ 
    \item Compute the global mean in the tangent space:
    \[
    \bar{Z} = \frac{1}{N} \sum_{i=1}^N Z_i.
    \]
    \item Compute the within-class scatter matrix:
    \[
    S_W = \sum_{c=1}^C \sum_{i : y_i = c} (Z_i - \bar{Z}_c)(Z_i - \bar{Z}_c)^\top.
    \]
    \item Compute the between-class scatter matrix:
    \[
    S_B = \sum_{c=1}^C N_c (\bar{Z}_c - \bar{Z})(\bar{Z}_c - \bar{Z})^\top.
    \]
    \item select the $d$ eigenvectors corresponding to the largest eigenvalues of 
    $ 
    S_B U = \lambda S_W U,
    $ 
    \item Project the data into the low-dimensional space:
    $ 
    z_i = U^\top Z_i.
    $ 
\end{enumerate}
\end{algorithm}

Riemannian LDA respects the intrinsic geometry of manifold-valued data while
leveraging label information to maximize class separability.  It generalizes Euclidean LDA and can be used for classification, visualization,
or manifold-based feature extraction.   
Tangent-space approximations work well when the manifold curvature is moderate, and
the method naturally extends to SPD, Grassmann, rotation groups, or other matrix manifolds.

\section{Riemannian Isomap}

Let $\{x_i\}_{i=1}^N \subset \mathcal{M}$ be points on a Riemannian manifold $(\mathcal{M}, g)$. 
Riemannian Isomap aims to compute low-dimensional embeddings $\{y_i\}_{i=1}^N \subset \mathbb{R}^d$ that preserve the manifold's intrinsic geometry.

\begin{enumerate}
    \item \emph{Neighborhood Graph:} Construct a $k$-nearest neighbor graph $\mathcal{G}$ using Riemannian distances
    \[
        \mathcal{N}(i) = \{ j \mid x_j \text{ is among the $k$ nearest neighbors of } x_i \text{ on } \mathcal{M} \}.
    \]
    \item \emph{Geodesic distance approximation:} Compute the shortest-path distances on the graph
    \[
        D_{ij} = \min_{\text{paths } x_i \to x_j} \sum_{(x_p,x_q) \in \text{path}} d_\mathcal{M}(x_p,x_q).
    \]
    \item  Compute low-dimensional embeddings $Y$ that preserve distances:
    \[
        Y = \arg\min_{Y \in \mathbb{R}^{N \times d}} \sum_{i<j} \big( \|y_i - y_j\| - D_{ij} \big)^2.
    \]
\end{enumerate}
The problem  can also be expressed as a trace optimization problem. 
Let $D^{(2)} \in \mathbb{R}^{N \times N}$ be the squared distance matrix with entries $D_{ij}^2$, and let 
\[
H = I_N - \frac{1}{N} \mathbf{1}\mathbf{1}^\top
\] 
be the centering matrix. The inner-product (Gram) matrix $B \in \mathbb{R}^{N \times N}$ is then
\[
B = -\frac{1}{2} H D^{(2)} H.
\]
The low-dimensional embedding $Y \in \mathbb{R}^{N \times d}$ is obtained by solving the trace maximization problem:
\begin{equation}
\begin{aligned}
\max_{Y \in \mathbb{R}^{N \times d}} & \quad \mathrm{tr}(Y^\top B Y) \\
\text{subject to} & \quad Y^\top \mathbf{1} = 0, \\
                   & \quad Y^\top Y = \Lambda_d,
\end{aligned}
\end{equation}
where $\Lambda_d$ contains the top $d$ eigenvalues of $B$. 
This ensures that the Euclidean embedding preserves the manifold's geodesic distances as well as possible.
\\
Riemannian Isomap is an extension of the classical Isomap algorithm to data lying on a Riemannian manifold $\mathcal{M}$. Unlike classical Isomap, which assumes Euclidean distances, Riemannian Isomap uses {geodesic distances} to better capture the intrinsic geometry of the manifold.
\\
Let $\mathcal{X} = \{x_1, \dots, x_N\} \subset \mathcal{M}$ be the dataset.
\\
For each point $x_i \in \mathcal{X}$, identify neighbors $\mathcal{N}(i)$ using either:
\begin{itemize}
    \item $k$-nearest neighbors
    \item $\epsilon$-radius neighborhood
\end{itemize}
 Connect neighbors with edges weighted by Riemannian distances:
\[
w_{ij} = 
\begin{cases} 
d_\mathcal{M}(x_i, x_j), & x_j \in \mathcal{N}(i) \\
\infty, & \text{otherwise}
\end{cases}
\]
where $d_\mathcal{M}(\cdot, \cdot)$ is the geodesic distance on $\mathcal{M}$.
\\
The second step consists in approximating the geodesic distance between all pairs of points using the shortest path on the neighborhood graph:
\[
D_\mathcal{M}(i,j) = \text{shortest-path distance between } x_i \text{ and } x_j
\]
Then we apply classical MDS to the geodesic distance matrix $D_\mathcal{M} = [D_\mathcal{M}(i,j)]$ to obtain a low-dimensional embedding $Y \in \mathbb{R}^{N \times d}$:
\[
H = I_N - \frac{1}{N} \mathbf{1}\mathbf{1}^\top, 
\quad 
B = -\frac{1}{2} H D_\mathcal{M}^{(2)} H
\]
where $D_\mathcal{M}^{(2)}$ contains squared geodesic distances, and $\Lambda_d$ contains the top $d$ eigenvalues of $B$. The columns of the matrix $Y$ are the $d$ eigenvectors 
corresponding to the $d$-largest  eigenvalues.  \\
The low-dimensional embedding can also be expressed as a trace maximization problem:
\[
\begin{aligned}
\max_{Y \in \mathbb{R}^{N \times d}} \quad & \mathrm{tr}(Y^\top B Y) \\
\text{s.t.} \quad & Y^\top \mathbf{1} = 0, \\
& Y^\top Y = \Lambda_d
\end{aligned}
\]
This ensures that the Euclidean embedding preserves the manifold's geodesic distances as accurately as possible.

\begin{algorithm}[H]
\caption{Riemannian Isomap}
\label{alg:riemannian-isomap}
\begin{algorithmic}[1]
\REQUIRE Data $\{x_i\}_{i=1}^N \subset \mathcal{M}$, target dimension $d$, neighborhood size $k$ or radius $\epsilon$
\ENSURE Low-dimensional embeddings $\{y_i\}_{i=1}^N \subset \mathbb{R}^d$

\STATE Construct neighborhood graph $\mathcal{N}(i)$ using $k$-NN or $\epsilon$-radius
\STATE Compute edge weights $w_{ij} = d_\mathcal{M}(x_i, x_j)$ for neighbors
\STATE Compute geodesic distances $D_\mathcal{M}(i,j)$ using shortest paths on the graph
\STATE Form squared distance matrix $D_\mathcal{M}^{(2)}$ and centering matrix $H = I_N - \frac{1}{N}\mathbf{1}\mathbf{1}^\top$
\STATE Compute Gram matrix $B = -\frac{1}{2} H D_\mathcal{M}^{(2)} H$
\STATE Eigen-decompose $B = V \Lambda V^\top$
\STATE Set $Y = V_d \Lambda_d^{1/2}$ using largest-$d$ eigenvectors
\RETURN $\{y_i\}_{i=1}^N$
\end{algorithmic}
\end{algorithm}

\begin{remark}
We can state the following important remarks:
\begin{itemize}
    \item The use of geodesic distances allows the method to respect the manifold's intrinsic geometry.
    \item Classical MDS ensures that the low-dimensional embedding preserves these distances as well as possible.
    \item For special manifolds such as spheres, SPD matrices, or Grassmann manifolds, geodesic distances and shortest paths must be computed respecting the manifold structure.
\end{itemize}
\end{remark}

\section{Riemannian support vector machine}

Support Vector Machines (SVMs) are large-margin classifiers originally defined in Euclidean spaces. 
When data lie on a Riemannian manifold $\mathcal{M}$, directly applying Euclidean SVM ignores the intrinsic geometry of the space. 
Riemannian Support Vector Machines (RSVMs) extend SVMs to manifold-valued data by exploiting tangent-space representations or Riemannian kernels.
\\
Let $\{(x_i, y_i)\}_{i=1}^N \subset \mathcal{M} \times \{-1,+1\}$ be labeled data points on a Riemannian manifold $(\mathcal{M}, g)$.
\\
A reference point $x_0 \in \mathcal{M}$ is chosen as the Fréchet mean:
\[
x_0 = \arg\min_{x \in \mathcal{M}} \sum_{i=1}^N d_{\mathcal{M}}^2(x, x_i),
\]
where $d_{\mathcal{M}}(\cdot,\cdot)$ denotes the geodesic distance.
Each data point is mapped to the tangent space at $x_0$ using the logarithmic map:
\[
z_i = \operatorname{Log}_{x_0}(x_i) \in T_{x_0}\mathcal{M}.
\]
\\
In the tangent space $T_{x_0}\mathcal{M}$, RSVM solves the standard soft-margin SVM problem:
\[
\begin{aligned}
\min_{w,\, b,\, Z} \quad &
\frac{1}{2} \|w\|^2 + C \sum_{i=1}^N Z_i \\
\text{subject to} \quad &
y_i \big( \langle w, z_i \rangle + b \big) \ge 1 - Z_i, \\
& Z_i \ge 0, \quad i = 1, \dots, N,
\end{aligned}
\]
where $w \in T_{x_0}\mathcal{M}$ is the normal vector of the separating hyperplane, $b \in \mathbb{R}$ is the bias term, $Z_i$ are slack variables, and $C > 0$ controls the trade-off between margin maximization and classification error.
\\
For a new point $x \in \mathcal{M}$, classification is performed via
\[
f(x) = \mathrm{sign}\!\left( \langle w, \operatorname{Log}_{x_0}(x) \rangle + b \right).
\]
This induces a decision boundary on $\mathcal{M}$ that corresponds to a geodesic hyperplane.
Alternatively, RSVM can be formulated intrinsically using Riemannian kernels based on geodesic distances:
\[
K(x_i, x_j) = \exp\!\left( -\frac{d_{\mathcal{M}}^2(x_i, x_j)}{2\sigma^2} \right).
\]
The dual optimization problem is then
\[
\begin{aligned}
\max_{\alpha} \quad &
\sum_{i=1}^N \alpha_i
- \frac{1}{2} \sum_{i,j=1}^N \alpha_i \alpha_j y_i y_j K(x_i, x_j) \\
\text{subject to} \quad &
0 \le \alpha_i \le C, \quad
\sum_{i=1}^N \alpha_i y_i = 0.
\end{aligned}
\]

\paragraph{Remarks.}
RSVM preserves the intrinsic geometry of the manifold by operating either in the tangent space or directly through geodesic-distance-based kernels. 
In Euclidean space, where $\mathcal{M} = \mathbb{R}^d$ and $\operatorname{Log}_{x_0}(x) = x - x_0$, RSVM reduces to the classical SVM formulation.

\section{Numerical experiments}
\label{sec:experiments}

We conduct comprehensive experiments to evaluate the proposed Riemannian dimensionality reduction methods against their Euclidean counterparts. This section describes the experimental setup, datasets, evaluation protocol, and presents detailed results.


\nd We evaluate our methods on   a diverse benchmark with real-world, manifold, and spherical datasets.



\nd Table~\ref{tab:datasets_benchmark} summarizes the characteristics of 12 benchmark datasets spanning three categories.

\begin{table}[htbp]
\centering
\caption{Benchmark dataset characteristics. $n$: samples, $d$: ambient dimension, $C$: classes.}
\label{tab:datasets_benchmark}
\begin{tabular}{llrrr}
\toprule
\textbf{Category} & \textbf{Dataset} & $n$ & $d$ & $C$ \\
\midrule
\multirow{4}{*}{\textit{Real-world}} 
& MNIST 8$\times$8   & 1797  & 64  & 10 \\
& Wine               & 178   & 13  & 3  \\
& Breast Cancer      & 569   & 30  & 2  \\
& Synthetic HD       & 600   & 50  & 4  \\
\midrule
\multirow{4}{*}{\textit{Manifold 3D}} 
& Swiss Roll         & 1000  & 100   & 4  \\
& S-Curve            & 1000  & 100   & 2  \\
& 3D Moons           & 600   & 100   & 2  \\
& 3D Circles         & 600   & 100   & 2  \\
\midrule
\multirow{4}{*}{\textit{Spherical 3D}} 
& Sphere Hard        & 600   & 100   & 4  \\
& Great Circle       & 600   & 100   & 4  \\
& Sphere Bands       & 600   & 100   & 3  \\
& Rings              & 400   & 100   & 2  \\
\bottomrule
\end{tabular}
\end{table}

\paragraph{Real-world datasets.} Standard classification benchmarks:
\begin{itemize} 
    \item \textbf{MNIST 8$\times$8}: Downsampled handwritten digits.
    \item \textbf{Wine}: Chemical analysis of wines from three cultivars.
    \item \textbf{Breast Cancer}: Wisconsin tumor diagnostic features.
    \item \textbf{Synthetic HD}: High-dimensional data with 10 informative and 10 redundant features among 50 total dimensions.
\end{itemize}

\paragraph{Manifold 3D datasets.} Nonlinear low-dimensional structures embedded in $\mathbb{R}^3$:
\begin{itemize}[noitemsep,topsep=0pt]
    \item \textbf{Swiss Roll}: 2D manifold rolled into 3D ($\sigma=0.5$).
    \item \textbf{S-Curve}: S-shaped surface ($\sigma=0.1$).
    \item \textbf{3D Moons}: Two interleaving half-circles with Gaussian noise.
    \item \textbf{3D Circles}: Concentric circles with scaling factor 0.5.
\end{itemize}

\paragraph{Spherical 3D datasets.} Data intrinsically lying on $\mathcal{S}^2$, designed to evaluate Riemannian methods:
\begin{itemize} 
    \item \textbf{Sphere Hard}: Four Fibonacci-distributed clusters with tangent noise ($\sigma=0.25$).
    \item \textbf{Great Circle}: Four arc segments along the equator ($\sigma=0.08$).
    \item \textbf{Sphere Bands}: Three latitudinal bands at $z \in \{-0.7, 0, 0.7\}$.
    \item \textbf{Rings}: Two interlocking orthogonal rings in $xy$ and $xz$ planes.
\end{itemize}

\begin{figure}[H]
    \centering
    \includegraphics[width=1.20\textwidth]{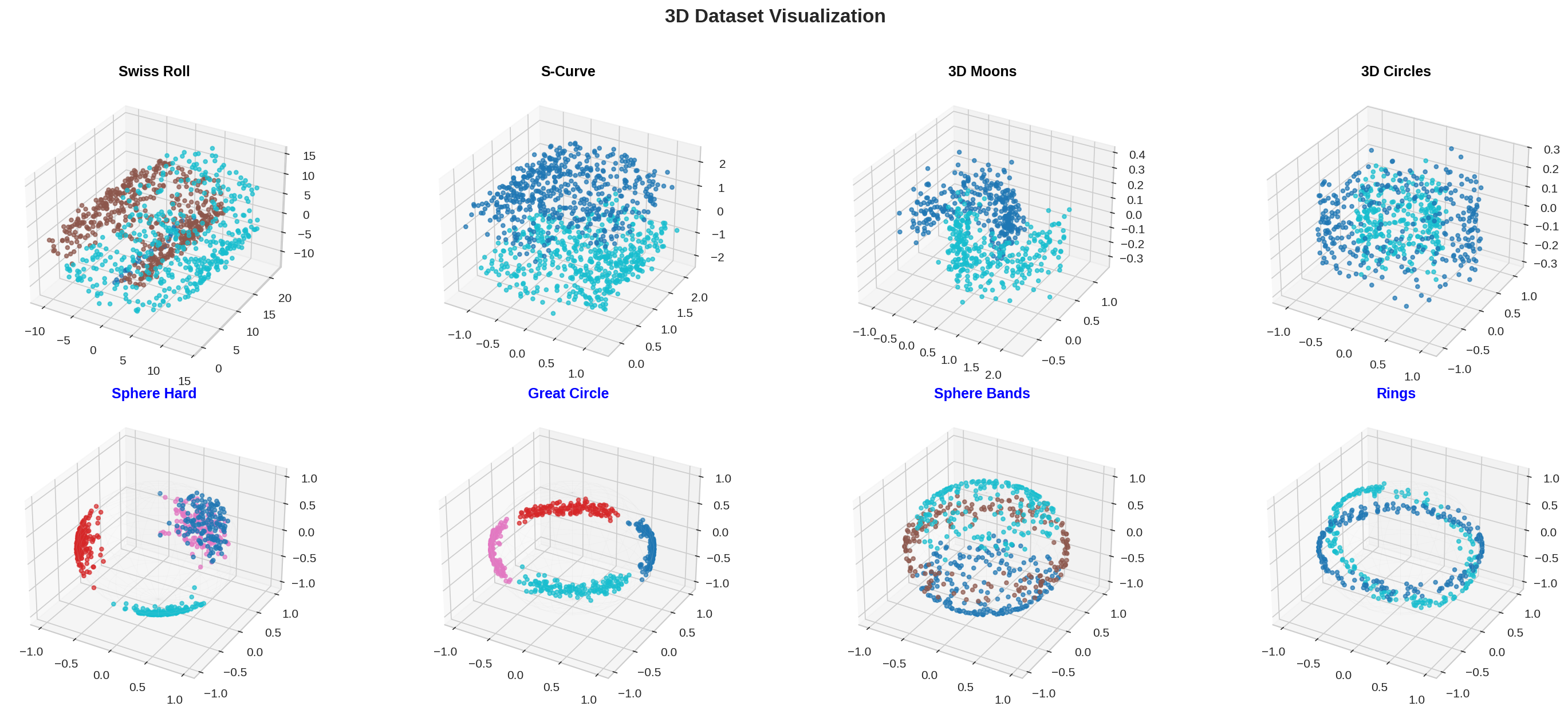}
    \caption{Visualization of 3D manifold and spherical datasets. Spherical datasets (bottom row) exhibit intrinsic geometry on $\mathcal{S}^2$.}
    \label{fig:datasets_3d}
\end{figure}



\nd Table~\ref{tab:hyperparams} summarizes the hyperparameter settings used throughout all experiments.

\begin{table}[H]
\centering
\caption{Hyperparameter configuration.}
\label{tab:hyperparams}
\begin{tabular}{ll}
\toprule
{Parameter} & {Value / Rule} \\
\midrule
Target dimension (benchmark) & $n_c = \min(3, d-1)$ \\
LDA components & $n_c^{\text{LDA}} = \min(n_c, C-1)$ \\
Number of neighbors & $k = \min(10, \max(3, \lfloor n/10 \rfloor))$ \\
KNN classifier & $k_{\text{KNN}} = 5$ \\
Train/test split & 70\% / 30\% (stratified) \\
Random seed & 42 \\
Fréchet mean tolerance & $\epsilon = 10^{-6}$ \\
Fréchet mean max iterations & 100 \\
R-RPCA ADMM iterations & 50 \\
\bottomrule
\end{tabular}
\end{table}


\nd We adopted the evaluation procedure:
\begin{enumerate}[noitemsep,topsep=0pt]
    \item Split data into 70\% training and 30\% test sets with stratified sampling.
    \item Fit the dimensionality reduction method on training data.
    \item Transform both training and test sets to the reduced space.
    \item Train a $k$-Nearest Neighbors classifier ($k=5$) on the embedded training data.
    \item Evaluate classification accuracy on the embedded test set.
\end{enumerate}

For transductive methods (R-LE, R-Isomap) that do not support out-of-sample extension, we concatenate train and test sets before embedding, then split the resulting coordinates for classification.

\nd We reported classification accuracy (\%) as the primary metric:
\begin{equation}
\text{Accuracy} = \frac{1}{|\mathcal{D}_{\text{test}}|} \sum_{i \in \mathcal{D}_{\text{test}}} \mathbf{1}[\hat{y}_i = y_i] \times 100
\end{equation}

\nd Experiments were conducted in Python 3.10 using scikit-learn~\cite{sklearn} for Euclidean baselines and custom implementations for Riemannian methods. All experiments use stratified subsampling with a maximum of 2000 samples per dataset to ensure computational tractability.

\emph{Results on benchmark  datasets}.  
Table~\ref{tab:benchmark_results} presents the classification accuracy on the 12 benchmark datasets with $n_c = 3$ components.

\begin{table*}[htbp]
\centering
\caption{Classification accuracy (\%) on benchmark datasets with $n_c = 3$ components. }
\label{tab:benchmark_results}
\resizebox{\textwidth}{!}{%
\begin{tabular}{ll|cccc|cccc|cccc}
\toprule
& & \multicolumn{4}{c|}{\textbf{Real-world}} & \multicolumn{4}{c|}{\textbf{Manifold 3D}} & \multicolumn{4}{c}{\textbf{Spherical 3D}} \\
\cmidrule(lr){3-6} \cmidrule(lr){7-10} \cmidrule(lr){11-14}
\textbf{Type} & \textbf{Method} & \rotatebox{60}{MNIST} & \rotatebox{60}{Wine} & \rotatebox{60}{Cancer} & \rotatebox{60}{Synth.} & \rotatebox{60}{Swiss} & \rotatebox{60}{S-Curve} & \rotatebox{60}{Moons} & \rotatebox{60}{Circles} & \rotatebox{60}{Sphere} & \rotatebox{60}{Circle} & \rotatebox{60}{Bands} & \rotatebox{60}{Rings} \\
\midrule
\multirow{4}{*}{\textit{Euclidean}} 
& PCA     & 73.5 & 98.2 & 93.0 & 50.6 & 74.0 & 90.0 & 84.4 & 80.0 & 98.3 & 78.9 & 74.4 & \textbf{92.5} \\
& LDA     & 85.6 & \textbf{100.0} & 96.5 & \textbf{86.1} & 94.3 & 97.3 & 87.2 & 48.9 & 99.4 & \textbf{100.0} & 97.8 & 73.3 \\
& Isomap  & 89.6 & 96.3 & 95.3 & 61.1 & \textbf{96.7} & \textbf{98.3} & 85.0 & \textbf{100.0} & \textbf{100.0} & \textbf{100.0} & 92.2 & 73.3 \\
\midrule
\multirow{6}{*}{\textit{Riemannian}} 
& R-PGA   & 72.0 & \textbf{100.0} & 93.6 & 54.4 & 78.0 & \textbf{98.3} & \textbf{90.0} & 57.8 & \textbf{100.0} & 99.4 & 97.8 & 96.7 \\
& R-RPCA  & 73.3 & \textbf{100.0} & 93.6 & 57.8 & 78.0 & \textbf{98.3} & \textbf{90.0} & 57.8 & \textbf{100.0} & 99.4 & 97.8 & 96.7 \\
& R-ONPP  & 41.3 & 94.4 & 93.0 & 40.0 & 77.3 & 98.0 & \textbf{90.0} & 56.1 & \textbf{100.0} & 99.4 & 97.2 & 95.8 \\
& R-LE    & \textbf{90.2} & 98.2 & 96.5 & 63.3 & 72.3 & 98.0 & 88.3 & 56.7 & \textbf{100.0} & \textbf{100.0} & 96.7 & 94.2 \\
& R-LDA   & 82.2 & \textbf{100.0} & \textbf{97.7} & 81.1 & 78.3 & \textbf{98.3} & 85.6 & 54.4 & \textbf{100.0} & \textbf{100.0} & 96.7 & 82.5 \\
& R-Isomap& 84.8 & \textbf{100.0} & 92.4 & 55.0 & 75.3 & 98.0 & 85.0 & 58.9 & \textbf{100.0} & \textbf{100.0} & 91.7 & 68.3 \\
\bottomrule
\end{tabular}%
}
\end{table*}

\nd \emph{Key observations.}
\begin{itemize}
    \item On \textbf{spherical datasets}, Riemannian methods achieve near-perfect accuracy (96--100\%), demonstrating the advantage of respecting the underlying manifold geometry.
    \item \textbf{R-LE} achieves the best result on MNIST (90.2\%), outperforming all Euclidean methods.
    \item \textbf{LDA/R-LDA} dominate on datasets with clear class structure (Wine, Breast Cancer).
    \item \textbf{Isomap} excels on manifold datasets (Swiss Roll, S-Curve, Circles).
\end{itemize}

\begin{figure}[H]
    \centering
    \includegraphics[scale=0.28]{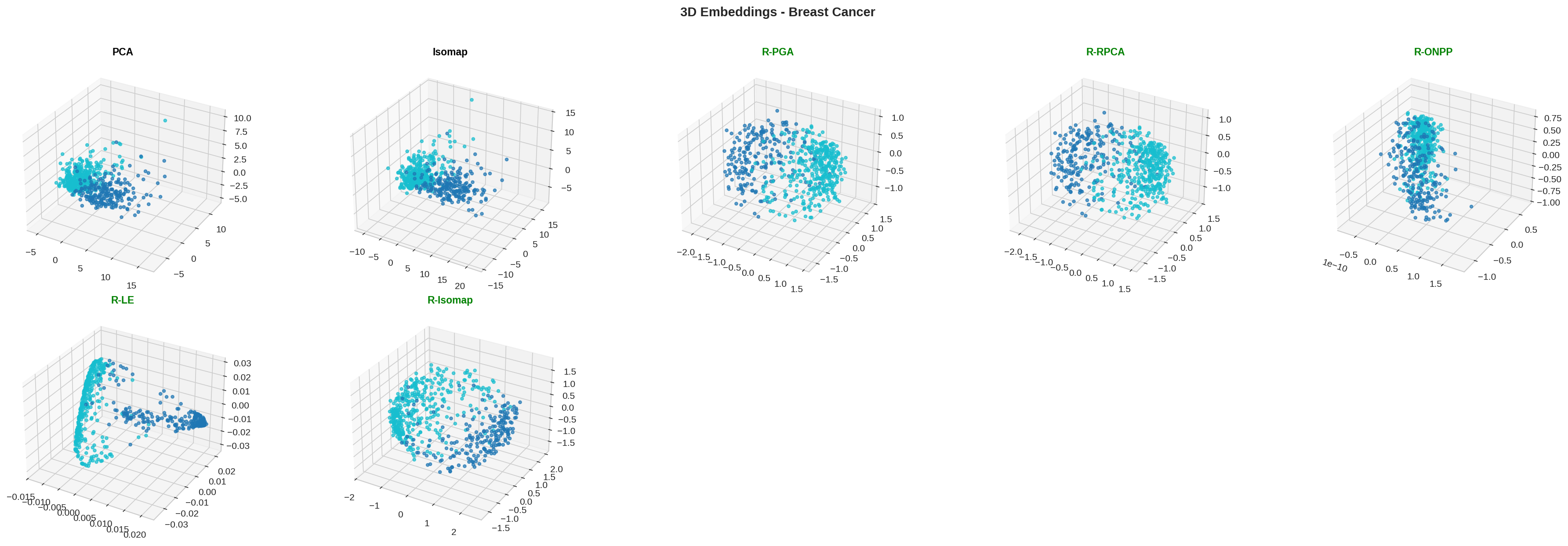}
    \caption{3D embeddings of Breast Cancer dataset.}
    \label{fig:emb3d_cancer}
\end{figure}

\begin{figure}[H]
    \centering
   \includegraphics[scale=0.30]{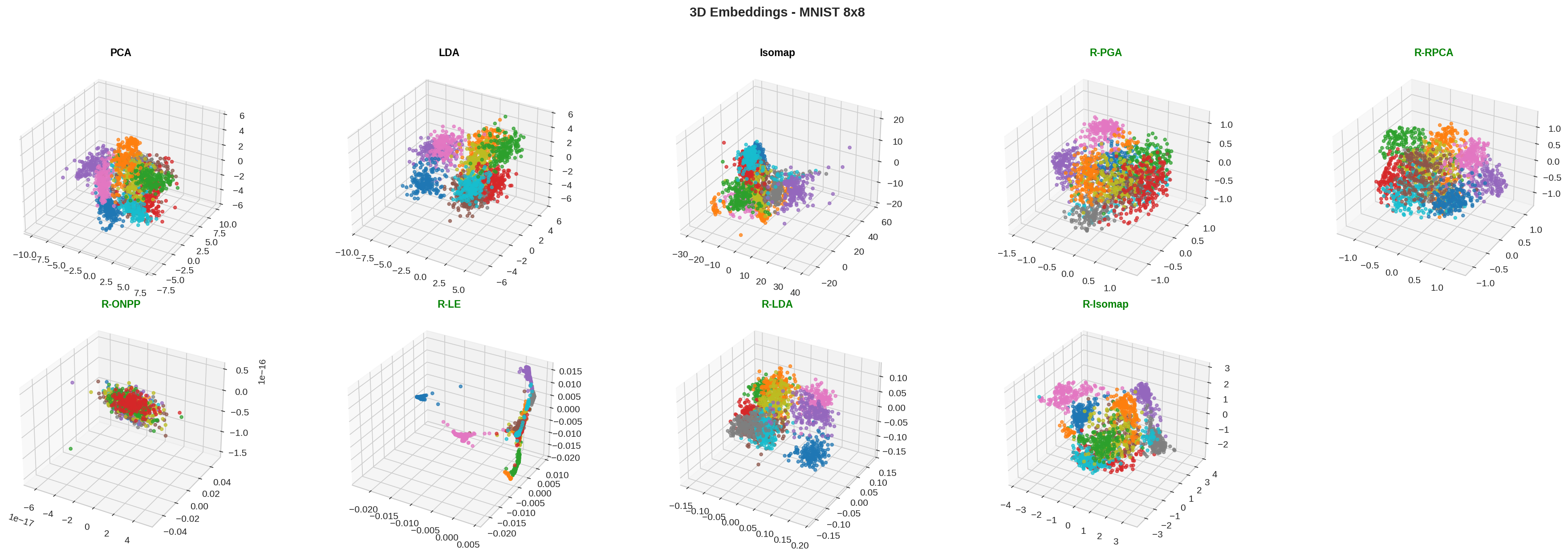}
    \caption{3D embeddings of MNIST dataset.}
    \label{fig:emb3d_mnist}
\end{figure}

\begin{figure}[H]
    \centering
    \includegraphics[scale=0.30]{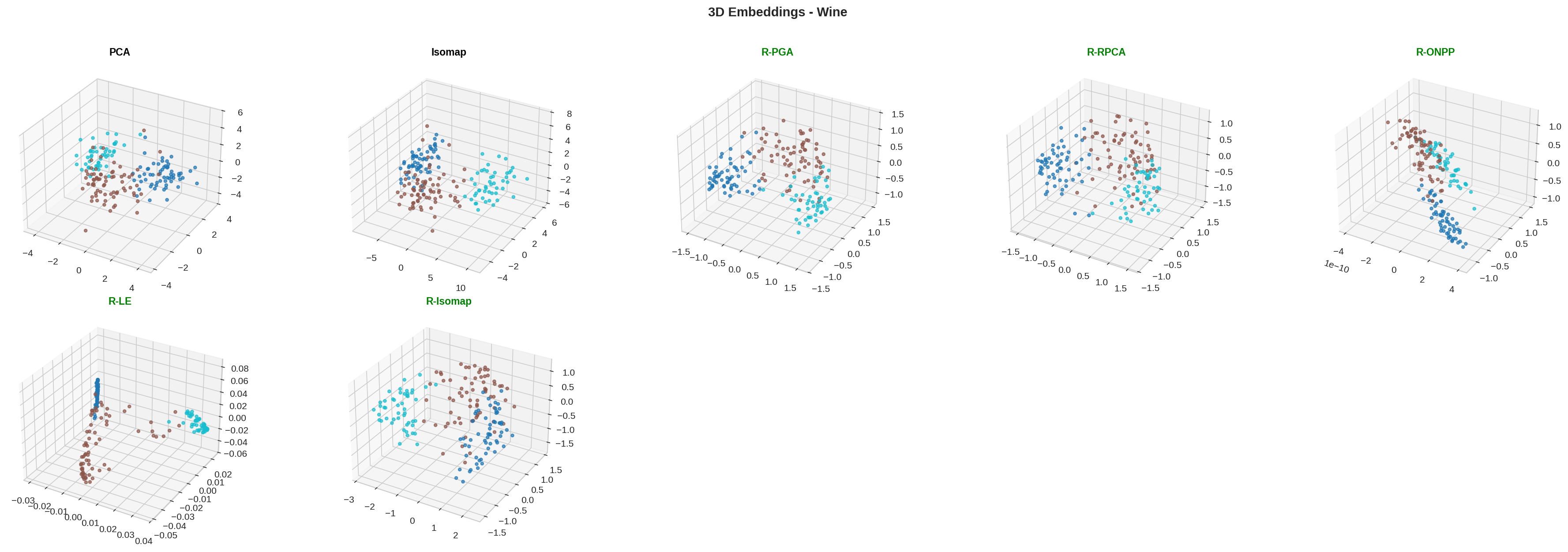}
    \caption{3D embeddings of Wine dataset. }
    \label{fig:emb3d_wine}
\end{figure}

\begin{figure}[H]
    \centering
    \includegraphics[scale=0.30]{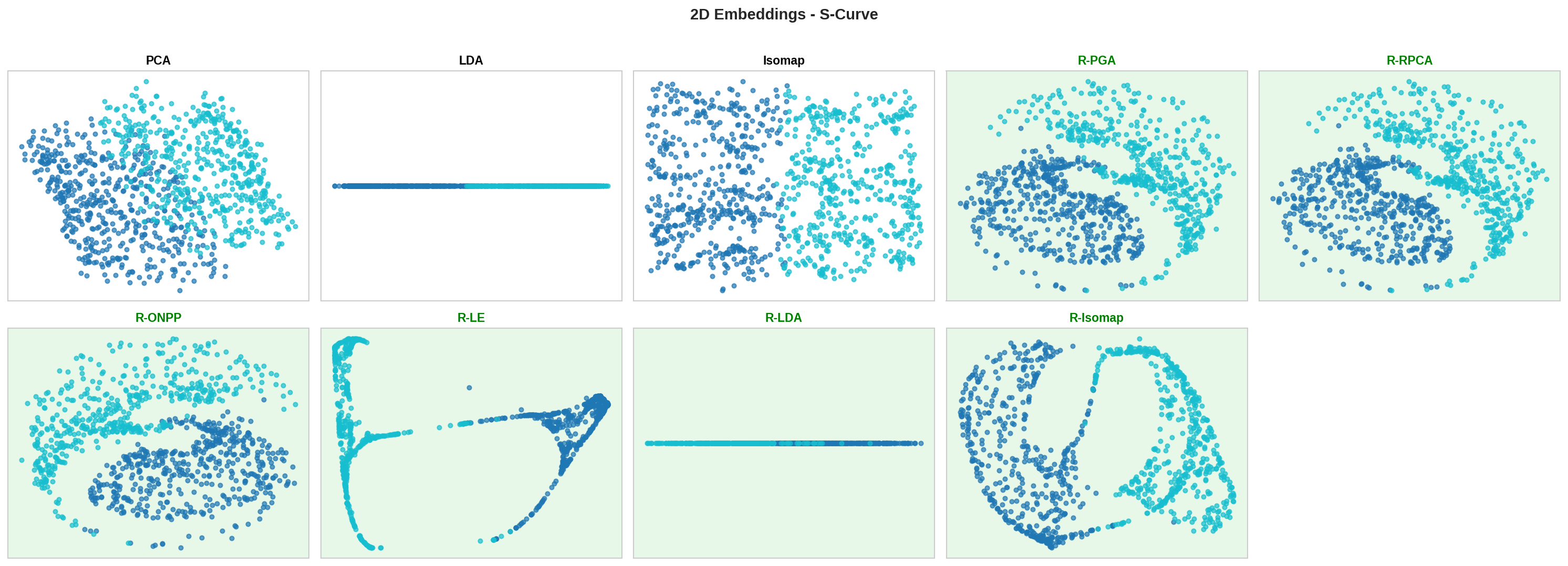}
    \caption{2D embeddings of S-Curve dataset}
    \label{fig:emb2d_scurve}
\end{figure}

\begin{figure}[h]
    \centering
    \includegraphics[width=1.15\textwidth]{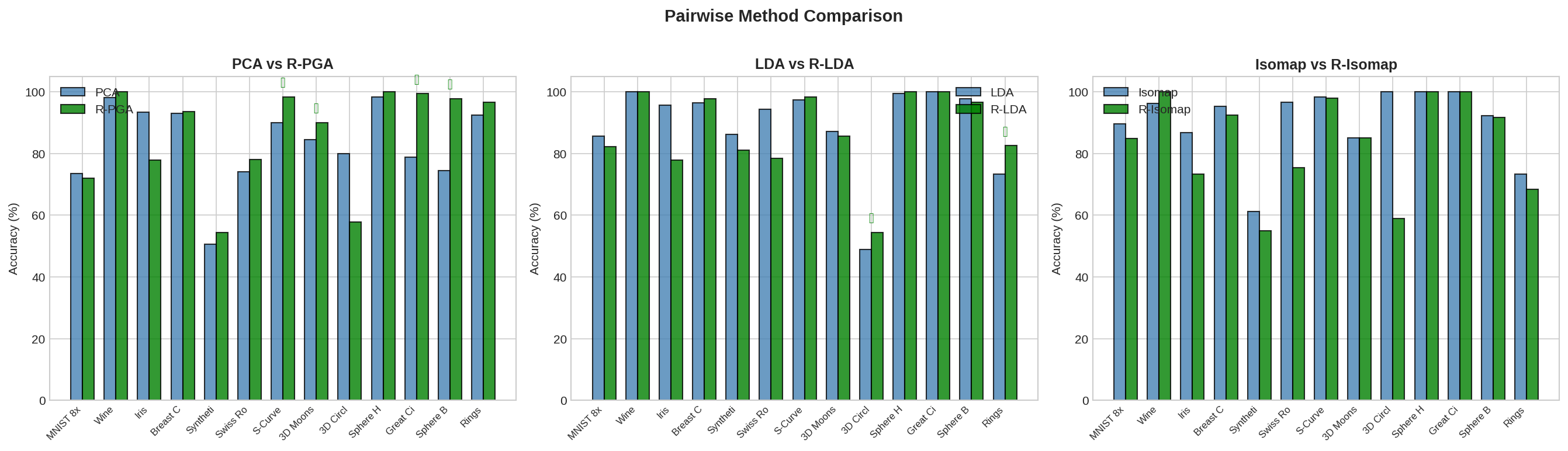}
    \caption{Pairwise comparison of Euclidean vs. Riemannian methods across benchmark datasets. }
    \label{fig:pairwise}
\end{figure}

\nd Figures~\ref{fig:emb3d_cancer}--\ref{fig:pairwise} illustrate the embedding quality and comparative performance of all methods.



\section{Conclusion}
In conclusion, dimensionality reduction on manifold-valued data represents a crucial advancement over classical Euclidean techniques, particularly for modern applications where data inherently reside on nonlinear spaces. By incorporating Riemannian geometry, methods such as Principal Geodesic Analysis, Riemannian discriminant analysis or Riemannian Laplacian Eigenmaps provide geometry-aware representations that more faithfully capture the intrinsic structure of the data. Compared to traditional linear approaches, these manifold-based techniques offer improved fidelity in low-dimensional embeddings and enhanced performance in downstream tasks such as classification. Our study reinforces the necessity of respecting the underlying geometry in data analysis and underscores the growing relevance of Riemannian approaches in machine learning, computer vision, and related fields where non-Euclidean data are prevalent.


\begin{thebibliography}{99}

\bibitem{absil2009optimization}
	P.-A. Absil, R.~Mahony, and R.~Sepulchre.
	 { Optimization Algorithms on Matrix Manifolds}.
	 Princeton University Press, 2009.

     \bibitem{arsigny2007geometric}
V.~Arsigny, P.~Fillard, X.~Pennec, and N.~Ayache,
 Geometric means in a novel vector space structure on symmetric positive-definite matrices,
{SIAM Journal on Matrix Analysis and Applications}, 29(1), 328--347 (2007).

\bibitem{carter2011statistical}
K.~M.~Carter and A.~O.~Hero III,
 Dimensionality reduction on statistical manifolds,
 Technical report, University of Michigan, 2011. 


     \bibitem{ElGuide2025}
M. El Guide, A. El Ichi, K. Jbilou, L. Rieched and H. Alqahtani, A Unified Trace-Optimization Framework  for Multidimensionality Reduction, arXiv preprint arXiv:2601.00729.

\bibitem{fletcher2004principal}
P.~T. Fletcher, C.~Lu, S.~M. Pizer, and S.~Joshi,
 Principal geodesic analysis for the study of nonlinear statistics of shape,
 {IEEE Transactions on Medical Imaging}, 23(8), 995--1005 (2004).

\bibitem{fukunaga2013introduction}
K.~Fukunaga,
 {Introduction to Statistical Pattern Recognition},
 Academic Press, 2nd ed., 2013.

 \bibitem{harandi2014manifold}
M.~T. Harandi, M.~Salzmann, and R.~Hartley,
 Manifold learning and classification on symmetric positive definite matrices,
 {IEEE Transactions on Pattern Analysis and Machine Intelligence}, 36(4), 767--782 (2014).

 \bibitem{Jbilou2026}
 K. Jbilou, Riemannian Manifold-Based Model Reduction for Large Nonlinear Dynamical Systems, submitted, 2026


\bibitem{jolliffe2002principal}
I.~T. Jolliffe,
 {Principal Component Analysis},
 Springer, New York, 2002.

\bibitem{larik2024rgda}
N.~A.~Larik, H.~Quan, and T.~Ji,
 Riemannian geodesic discriminant analysis–minimum Riemannian mean distance for image set classification,
 {Mathematics}, 12(14):2164, 2024. 


\bibitem{lin2006riemannian}
T.~Lin, H.~Zha, and S.~U.~Lee,
  Riemannian manifold learning for nonlinear dimensionality reduction,
 in {Lecture Notes in Computer Science}, vol. 3951, pp. 44--55, Springer, 2006.
 
 \bibitem{sklearn}
 	F.~Pedregosa, G.~Varoquaux, A.~Gramfort, et~al.,
 	Scikit-learn: Machine learning in {P}ython,
 	{Journal of Machine Learning Research}, 12:2825--2830, 2011.
 

\bibitem{pennec2006intrinsic}
X.~Pennec, Intrinsic statistics on Riemannian manifolds,
 {Journal of Mathematical Imaging and Vision}, 25(1), 127--154, 2006.

 \bibitem{rodriguez2025rpca}
O.~Rodríguez,
 Riemannian principal component analysis,
 {arXiv preprint}, 2025.

 \bibitem{roweis2000nonlinear}
S.~T. Roweis and L.~K. Saul,
 Nonlinear dimensionality reduction by locally linear embedding,
 {Science}, 290(5500), 2323--2326 (2000).



\bibitem{pennec2019riemannian}
X.~Pennec, S.~Sommer, and P.~T. Fletcher,
 {Riemannian Geometry and Statistical Learning},
 Academic Press, 2019.

 \bibitem{tan2019smooth}
T.~Tan, et al.,
 A nonlinear dimensionality reduction framework using smooth geodesics,
 {Pattern Recognition}, 87:226–236, 2019. 


\bibitem{tenenbaum2000global}
J.~B. Tenenbaum, V.~de Silva, and J.~C. Langford,
 A global geometric framework for nonlinear dimensionality reduction,
 {Science}, 290(5500), 2319--2323 (2000).


 \bibitem{xu2023camel}
N.~Xu and Y.~Liu,
CAMEL: Curvature-augmented manifold embedding and learning, {arXiv preprint}, 2303.02561, 2023.

\bibitem{yang2020nested}
C.-H.~Yang and B.~C.~Vemuri,
 Nested Grassmannians for dimensionality reduction with applications,
 {Journal of Machine Learning for Biomedical Imaging}, 1(2022):1--21, 2022. 


\bibitem{zhang2019mixture}
Y.~Zhang, J.~Xing, and M.~Zhang,
 Mixture probabilistic principal geodesic analysis,
 {arXiv preprint}, 1909.01412, 2019. 
 
\bibitem{zhu2018rmml}
P.~Zhu, H.~Cheng, Q.~Hu, Q.~Wang, and C.~Zhang,
 Towards generalized and efficient metric learning on Riemannian manifolds,
in {Proc. IJCAI}, pp. 3235–3241, 2018.

\end{thebibliography}
\end{document}